\documentclass[ letterpaper,10 pt,conference ]{IEEEtran}  
\IEEEoverridecommandlockouts 

\usepackage{times}
\usepackage{epsfig}
\usepackage{graphicx}
\usepackage{amsmath}
\usepackage{amssymb}

\usepackage[]{algorithm2e}
\usepackage{algpseudocode}
\usepackage{amsmath,amssymb}
\usepackage{caption}
\usepackage{subcaption}

\usepackage{times}
\graphicspath{{figures/}}

\usepackage[pagebackref=true,breaklinks=true,letterpaper=true,colorlinks,bookmarks=false]{hyperref}
\usepackage{color}
\usepackage{bbm}

\newcommand{\hmay}{{\bf H}}

\newcommand{\uu}{{\bf u}}

\newcommand{\ww}{{\bf  w}}

\definecolor{CommentUrgent}{rgb}{0.7,0,0}
\definecolor{CommentPA}{rgb}{0.0,0.7,0.0} 
\definecolor{CommentLP}{rgb}{0.6,0.3,0.6} 
\definecolor{CommentPP}{rgb} {0.0,0.0,0.7} 

\renewcommand{\baselinestretch}{0.97}
\begin{document}

 \title{Geometric Multi-Model Fitting with a Convex Relaxation Algorithm}

\author{Paul Amayo, Pedro Pini\'es, Lina M. Paz, Paul Newman \and
\thanks{Authors are from the Oxford Robotics Institute, Dept. Engineering Science, University of Oxford, UK. {\tt\small \{pamayo, ppinies, linapaz, pnewman\}@robots.ox.ac.uk} } }

\maketitle

\maketitle

\begin{abstract}

We propose a novel method to fit and segment multi-structural data via convex relaxation. Unlike greedy methods --which maximise the number of inliers-- this approach efficiently searches for a soft assignment of points to models by minimising the energy of the overall classification. Our approach is similar to state-of-the-art energy minimisation techniques which use a global energy. However, we deal with the scaling factor (as the number of models increases) of the original combinatorial problem by relaxing the solution. This relaxation brings two advantages: first, by operating in the continuous domain we can parallelize the calculations. Second, it allows for the use of different metrics which results in a more general formulation.

We demonstrate the versatility of our technique on two different problems of estimating structure from images: plane extraction from RGB-D data and homography estimation from pairs of images. In both cases, we report accurate results on publicly available datasets, in most of the cases outperforming the state-of-the-art.
\end{abstract}

\section{Introduction}
Many important tasks in computer vision such as homography estimation, plane detection and motion estimation demand the ability to fit geometric models onto noisy data. This is a non-trivial task given that the scene typically consists of multiple geometric structures. Moreover, the observed data is likely contaminated with noise from different sources including measurement sensor noise and outliers. These are the main factors leading to biased solutions. Therefore many multi-model fitting algorithms are driven fundamentally by their capacity to robustly deal with the complexity of the data and unknown distributions of data errors and outliers \cite{pham2014interacting}. 
These extracted geometric models are often drivers of other algorithms such as robotic navigation, dense reconstruction and multi-object tracking; applications in which time performance that is close to real-time is actively sought. This provides a need for  multi-model fitting algorithms that are not only robust to contamination but exhibit fast and repeatable runtimes.

The most common solution to the geometric multi-model fitting problem is RANSAC \cite{fischler1981random}. A greedy approach consisting of two steps: in the first step, a set of model proposals is sampled from the model parameter space in a hypothesis-verification fashion.  A refinement step is applied over the best-selected model supported by the maximum number of inliers. To deal with multiple models, an extended solution like \cite{torr1998geometric} suggests to apply RANSAC sequentially over the rest of data points. However, the optimality of the solution is not guaranteed therefore restricting its use in the presence of cluttered data. Running alongside these methods are greedy clustering methods that maximise the number of inliers to models \cite{toldo2008robust,magri2014t,magri2016multiple}.

Global energy based approaches \cite{isack2012energy, yu2011global, delong2012fast}  have gained popularity by presenting a more general optimisation framework that jointly fit all models present in multi-structural data.  In practice, global energy methods aim to find an optimal fitting solution by accounting for the model error in a data fidelity term for a given metric. In addition, the set of solutions can be constrained by encouraging spatial smoothness in a regularisation term. A more complete formulation considers also the number of models as a parameter to optimise for \cite{isack2012energy}. These approaches have shown to outperform greedy methods such as RANSAC at the expense of intractable computational complexity with respect to the number of models fitted. 

\begin{figure}[t]
\begin{center}
  \includegraphics[width=0.9\linewidth]{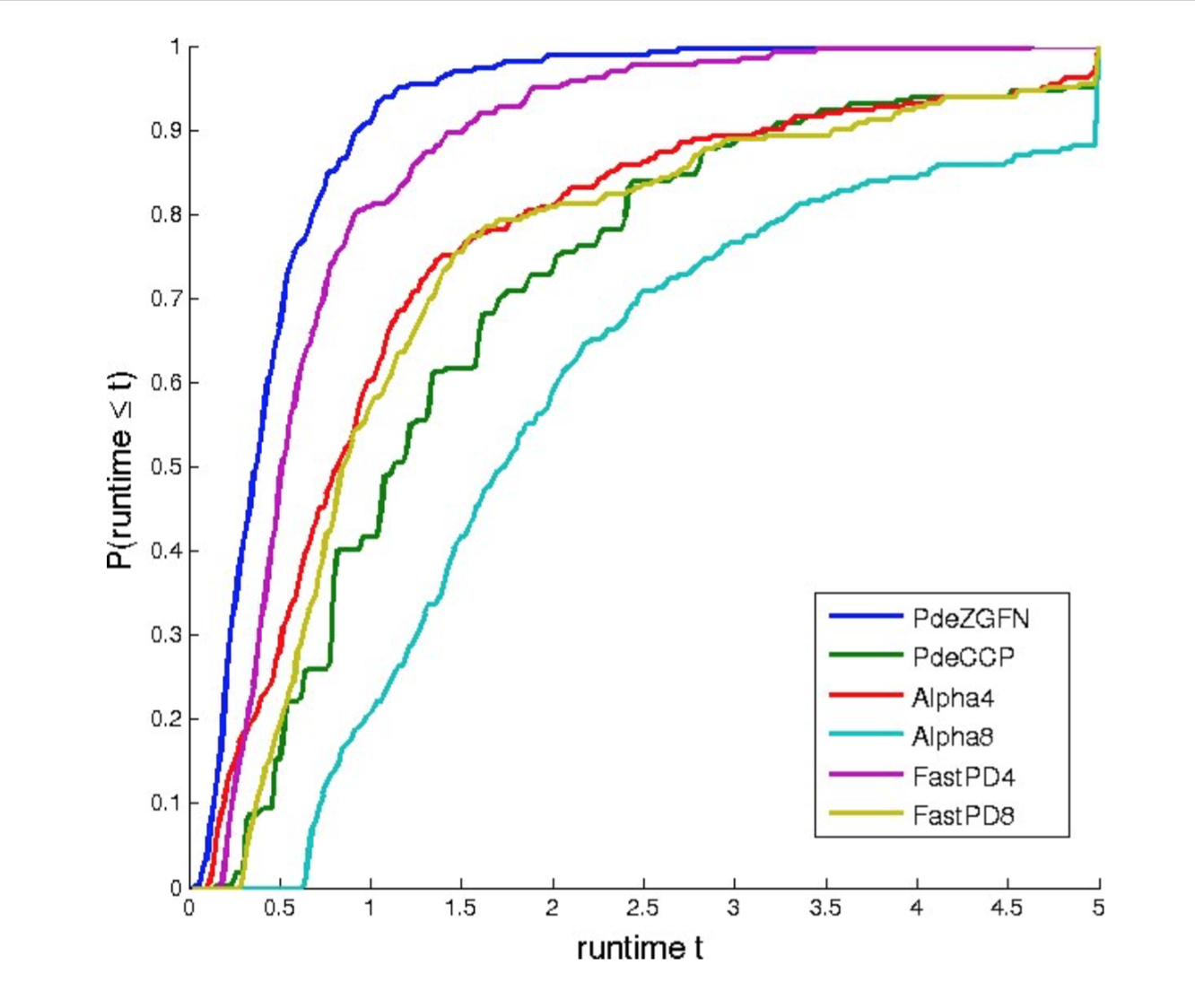}
\end{center}
  \caption{Statistical comparison of the runtimes of continuous optimisation approaches (PDEZGFN \cite{zach2008fast} and PDECCP \cite{chambolle2008convex}) against discrete approaches (FASTPD \cite{komodakis2007fast} and Alpha \cite{boykov2001fast}) for interactive image segmentation over all the images of the Graz dataset \cite{santner2010interactive}. The plot shows the cumulative distribution function $P(runtime<t)$ with PDEZGFN the fastest performing algorithm. Figure from \cite{nieuwenhuis-et-al-ijcv13}.}
\label{fig:runtime}
\end{figure}

Minimisation of this energy can be performed via combinatorial algorithms such as $\alpha$-expansion \cite{boykov2001fast} which operate in a discrete domain. These approaches have been largely applied to the multi-label image segmentation problem formalized as a graph cut Markov Random Field (MRF) based approach. In contrast, we present a general-purpose solution that exploits continuous multi-label optimisation, a framework that has been shown to be superior in terms of parallelisation and runtime performance \cite{nieuwenhuis-et-al-ijcv13}. Figure \ref{fig:runtime} shows the time performance of two continuous optimisation algorithms that utilise Partial Differential Equations(PDEs): PDEZGFN \cite{zach2008fast} and PDECCP \cite{chambolle2008convex}. Compared to their discrete counterparts, FASTPD \cite{komodakis2007fast} and $\alpha$-expansion \cite{boykov2001fast} with 4 and 8 connectivity respectively on a multi-label image segmentation  problem. From the Figure it can be seen that PDEZGFN exhibited the best run-time performance for the image-segmentation task.

Additionally \cite{nieuwenhuis-et-al-ijcv13} highlights an important characteristic of discrete methods: their tendency to have a high run-time variance among scenes with the same number of labels. To select the adequate solution for a particular application, $\alpha$-expansion solves a number of max flow problems until convergence. This solution, however highly depends not only on the input data but also the chosen label order. Moreover, the number of expansion steps per each max flow problem depend on the graph structure. These parameters strongly differ with the current labelling, leading to high variation during running time. In contrast, continuous approaches -- as the one we suggest here -- carry out the same computation steps on each data point reporting a smaller running time variance. In addition, the number of iterations until convergence does not depend on the initial label order.

Analogous to PEARL \cite{isack2012energy}, we convert the problem of geometric multi-model fitting to a multi-labelling problem where each model is represented through a label function over the data points. By relaxing the discrete labels we reduce the impact of the scaling factor which harms the original combinatorial problem. The optimal solution to the relaxed problem is obtained efficiently through a primal-dual solution \cite{chambolle2011first}. This is an important advantage since it supports point-wise updates, thus it facilitates a straightforward implementation on GPGPU hardware which in turns leads to a faster implementation as alluded to in \cite{nieuwenhuis-et-al-ijcv13}.

We summarise the main contributions in this paper as follows: 
\begin{itemize}
\item We propose a novel global energy-based approach to fit and segment multi-structural data via a convex relaxation algorithm, CORAL. Unlike greedy methods, which maximise the number of inliers, this approach efficiently searches for a soft assignment of points to models by minimising the energy of the overall classification. 
\item We provide a designed energy functional that encompasses spatial regularisation on a continuous label function while simultaneously minimising the number of labels that better adapt to the multi-structural data. It does not make any assumption on the sparsity of the data.
\item We demonstrate the adaptability of the approach to two important vision applications including multi-homography estimation and plane detection from RGB-D images. These applications show the versatility of our approach to different metrics with different norms used in the two instances.
\end{itemize}

We show that, when applied on both simulated and publicly available real datasets, the proposed approach outperforms other global energy based methods for geometric multi-model fitting. The paper is organised as follows. In Section \ref{sec:energy_fitting} we describe our framework on a continuous energy formulation. Section \ref{sec:convex_relaxation} describes the two-stage convex relaxation algorithm. The practical advantages of our geometric model fitting framework are demonstrated in Section \ref{sec:experiments_applications}. Finally, we discuss the advantages of our method and draw conclusions in Section \ref{sec:conclusions}.

\section{Energy Minimisation for Multi-model fitting}

\label{sec:energy_fitting}
We frame the geometric multi-model fitting as an optimisation problem where the quality of the solution is linked to an energy functional. The first term of the energy considers the geometric error of all data points to their corresponding models/labels. The energy also encompasses prior knowledge about locality in the solution by enforcing data points that are spatially close to have a higher likelihood of belonging to the same model. Finally, a different sort of regularisation that promotes compactness is imposed by favouring solutions that explain the data using as few models as possible. Equation \ref{eq:energy} represents these three ideas in a general formulation.
\begin{align}\label{eq:energy}
    \sum_{l=1}^L \left( \int_\Omega \rho_l(\uu,\phi_l(\uu)) + \lambda \omega_{\mathcal{N}} R( \nabla_{\mathcal{N}} \phi_l(\uu))d\Omega\right) +\beta L 
\end{align}
We refer to the first term $\rho_l$ in equation \ref{eq:energy} as a data fidelity term, defined over the data points $\uu \in  \Omega $, where $\Omega \subset \mathbb{R}^m$ represents a continuous domain. The data term can be seen as a geometric cost of a data point supporting a particular model.

The assignment of data points to their respective models is encapsulated through an indicator function
\begin{equation}\label{eq:indicator}
\phi_l(\uu) = 
 \begin{cases} 
   1 &  \uu \in L_l\\
   0 & \text{otherwise }
  \end{cases}
\end{equation}
where the uniqueness in the label assignment can be achieved by adding the constraint $\sum_{l=1}^L \phi_l(\uu) = 1$.

In practice, some data points might not be explained by a geometric model. For this case, we add a special label $\emptyset$, representing the outlier model. In this way, we can assign a constant cost $\gamma$ to points that cannot be explained by any geometric model. The model cost for the outlier model in simply given by  $\rho_\emptyset(\uu,\phi_\emptyset(\uu))=\gamma$.

The second term in Equation \ref{eq:energy} takes into account locality by promoting a homogeneous assignment of labels to neighbouring points. The $\nabla_{\mathcal{N}}$ operator calculates the gradient of the indicator function over the neighbourhood $\mathcal{N}$ of a point. The function $R$ is designed to penalise points that belong to the same neighbourhood but do not share the same model. This penalty can be evaluated using different norms $| \cdot |_{p}$. For instance, the standard 4-connected lattice implemented in graph cut algorithms \cite{boykov2001fast} can be obtained by implementing $\nabla_{\mathcal{N}}$ using forward differences combined with a Manhattan norm  $| \cdot |_{1,1}$ \footnote{In the 2D case, $|\nabla_{\mathcal{N}} \phi_l(\uu)|_{1,1} = |\nabla_{x}\phi_l(\uu)|+ |\nabla_{y}\phi_l(\uu)|$}. Examples of how to calculate $\nabla_{\mathcal{N}}$ and other norms for different applications can be found in the supplementary material. The parameter $\lambda$ controls the trade-off between the smoothness/locality cost and the model cost. The weights $\omega_{\mathcal{N}}$ allow us to have a finer control of the dependencies between neighbouring pixels. For example, in cases where the $k$ nearest neighbours for a given point are actually at a far distance, we reduce the effect of the smoothness term by decreasing the value of $\omega$ according to the separation. Finally, the third term in Equation \ref{eq:energy} penalises the number of models $L$ by adding a constant cost $\beta$ per model.

\subsection{Label Relaxation}
\label{sec:convex_relaxation}

The constraint in Equation \ref{eq:indicator} makes the problem combinatorial and NP-hard so it can only be approximately solved. We use a known fast relaxation approach \cite{zach2008fast} that transforms the original problem into a convex one. While this relaxation is not the tightest, it produces good results in practice. The relaxation is based on allowing $\phi_l(\uu)$ to take values in the interval $\phi_l(\uu) \in [0,1]$ instead of the binary set $\{0,1\}$. As a result, for a fixed number of models L, Equation (\ref{eq:energy}) with the linear equality constraint in $\phi_l(\uu)$ becomes a convex optimisation problem.

\subsection{Continuous optimisation algorithm }

The CORAL optimisation is shown in Algorithm \ref{alg:contopt}. Our approach exploits continuous optimisation in a two-stage iterative solution: First, an inner primal-dual iterative algorithm is carried out to impose spatial regularisation. Second, we employ an outer iteration to minimise for the number of models.
\begin{algorithm}[t]
\{Initialisation\}\;
 Propose $s$ models using RANSAC\;
 \While{not converged}{
  Primal Dual Optimisation\;
  Merge Models\;
  Re-estimate models\;  
 }
 \caption{CORAL}
 \label{alg:contopt}
\end{algorithm}

The solution adopted to solve the energy in Equation (\ref{eq:energy}) depends on the selection of the functionals $\rho_l$ and $R$. Non-smooth norms such as the $L_1$ norm have shown to be robust to reject outliers. However, its non-differentiable intrinsics prevents us of using standard optimisation techniques. Recent achievements on continuous optimisation \cite{chambolle2011first} show that non-smooth priors used in similar relaxed convex problems can be transformed into saddle point problems and then a first order primal-dual algorithm exists to find the optimal solution. In this paper, we apply the same transformation to our relaxed constrained problem.  
\begin{align}\label{eq:legendre-fenchel}
 \int_{\Omega} |\nabla \phi_l(\uu)|_{p} d\Omega \ = \ & \max_{\Psi_l(\uu)} \int_{\Omega} \nabla \phi_l(\uu) \cdot \Psi_l(\uu) d\Omega \\
                                         & s.t. \ |\Psi_l(\uu)|_{p *} \leq 1 \label{eq:legendre-fenchel-const}
\end{align}
where $\Psi_l(\uu): \Omega \rightarrow \mathbb{R}^2$ is known as the dual function of $\phi_l(\uu)$ and $|\cdot|_{p}$ and $|\cdot|_{p *}$ are dual norms. Although this transformation seems to apparently increase the complexity, the counterpart is that we can now use well known first order methods available for smooth problems to find the global solution of the relaxed energy.

The main steps for the global energy optimisation are summarised as follows:

\begin{algorithm}[t]
Initialisation\;
$\tau,\alpha>0, \theta\in[0,1]$\;
$\phi^0= \bar{\phi}^0=\Psi^0=0 $\;
\While{k $<$ N}{
{Dual Step}\;
$\Psi^{k+1}=\pi_{\Psi}(\Psi^k +\tau\nabla\bar{\phi}^k)$\;
{Primal Step}\;
$\phi^{k+1}=\pi_{\phi}(\phi^k-\alpha(\rho_l(\uu,\phi^k)+\lambda\nabla^T \Psi^{k+1})$\;
{Relaxation step}\;
$\bar{\phi}^{k+1}=\phi^{k+1}+\theta(\phi^{k+1}-\phi^k)$
}
\caption{Primal Dual Optimisation}
\label{alg:PrimalDual}
\end{algorithm}

\begin{enumerate}
\item Geometric error and smoothing optimisation:  
We summarise the first order primal dual optimisation  \cite{chambolle2011first} for the proposed energy in algorithm  \ref{alg:PrimalDual}. 
    
Algorithm \ref{alg:PrimalDual} shows the steps of the primal dual optimisation where $\tau,\alpha$ are the step size parameters of the algorithm, with $\theta$ controlling the relaxation.  The values of these are determined using the diagonal preconditioning scheme \cite{pock2011diagonal}.

To fulfil the constraint in Equation \ref{eq:legendre-fenchel-const}, the gradient ascent step of the dual variable gets projected onto the feasible set with the projection $\pi_{\Psi}(\cdot)$ defined by
\begin{equation}
\pi_{\Psi}(\Psi)=\frac{\Psi}{\max(1,|\Psi|_{p*})}
\end{equation}
Similarly, the function $\pi_{\phi}(\cdot)$ projects the gradient descent step of the primal variable onto the simplex $\phi_l(\uu) \in [0,1] \ s.t. \ \sum_{l=1}^L \phi_l(\uu) = 1$. The projection onto the simplex is explained in the supplementary material.    

\item Models reduction: 
Regularisation for compactness of the solution is imposed by adding a fixed cost multiplied to the number of models in Equation \ref{eq:energy}. Although it is not explicitly formulated into the primal dual optimisation, the smoothing optimisation has the unexpected benefit of sometimes reducing the label compactness energy. The smoothing optimisation reduces the smoothness energy by enforcing a single label for spatially connected regions. Then redundant models are merged in this way leading to a more compact solution. 

This however only holds when two models are spatially connected. If two models with similar parameters that are not spatially connected are merged into a single model this does not reduce the smoothness energy. To account for this case an extra step is performed after the primal dual optimisation that merges separated models with similar parameters. In the presence of noise, merging two models results in an increase in the geometric error energy.  If this increase is however less than $\beta$, the global energy still decreases. This extra model merging step explicitly performs an optimisation for the number of models in the solution, ensuring an optimal compact solution. 
\end{enumerate}

Analogous to the $\alpha$-expansion algorithm \cite{isack2012energy}, our proposed optimisation cannot be directly applied to continuous data as the number of possible labels for a model with $p$ parameters is $\mathbb{R}^p$. To reduce the search space, stochastic sampling for a finite number of models $s$ should be performed. Moreover, to re-estimate models after the primal-dual optimisation is completed we need a one-to-one correspondence between indicator functions and models. To this end, we threshold the continuous labels by selecting the maximum value at each point. The geometric error can be further reduced by re-estimating models based on the assignment given by the thresholded solution. This provides a new initialisation point to the inner primal dual optimisation.

\section{Experiments and Applications}
\label{sec:experiments_applications}
 In this section we employ our proposed algorithm to address the extraction of planar regions in images for two different problems: the first corresponds to the known multi-homography estimation from two views. The second, is plane detection from a single RGB-D image. Our motivation is driven by the fact that detection of geometric structure from images is of widespread importance to applications such as camera calibration, camera motion estimation and surface reconstruction. We argue that existing approaches are limited for real-time functionality.  
 
 The implicit assumption in both cases is that scenes consist mostly of man-made objects commonly found in urban outdoor and indoor environments (e.g. buildings, walls, screens, desks, etc.). The image setup choice (i.e. colour images versus range data) allows us to deal with different situations. On the one hand, indoor environments usually expose texture-less surfaces. In addition, matching algorithms will be affected by low-light conditions thus motivating the use of an RGB-D sensor. On the other hand, outdoor scenes are more abundant in texture. However, the use of a depth sensor is limited by its practical maximum range. A two-view homography estimation approach for plane detection with feature correspondences would result in a better choice. Under these assumptions, we show in the next sections the versatility of the proposed formulation with no special distinction on the distribution of data over the space domain. Furthermore we show the ability of our formulation to handle the different norms best suited for the two different situations.

\subsection{Two-view multi-homography estimation}
Our first application considers two views of a static scene with multiple planes. Given an sparse set of $n$ pixel correspondences in homogeneous coordinates between the two views $\uu_i = (\uu^1_i, \uu^2_i) \in \mathbb{R}^2,\ i=1 \cdots n$, the homography $\hmay_{21} \in \mathbb{R}^{3\times3}$ establishes the mapping of pixels from the first view to the second view through an observed plane. This operation is denoted by $\uu_i^{2} = \hmay_{21} \uu_i^{1}$. We aim to find the classification of pixel matches with respect to several homographies while simultaneously rejecting outliers. A simplified version of the energy in Equation \ref{eq:energy} for this problem is proposed in Equation \ref{eq:homography_energy},
\begin{align}\label{eq:homography_energy}
    & \sum_{l=1}^L\bigg(\frac{1}{2}\sum_{i=1}^n (\| D(\uu_i, \hmay_{12}) \|_{\Sigma_{12}} +  \| D(\uu_i, \hmay_{21}) \|_{\Sigma_{21}}) \phi_l(\uu) \nonumber \\
    & + \lambda \sum_{i=1}^n \omega_{\mathcal{N}} |\nabla_{\mathcal{N}} \phi_l(\uu) |_{1,1} \bigg)  + \beta L 
\end{align} 
The data term is chosen to account for the symmetric transfer error of the re-projection operation. \footnote{Here, we refer to $D$ as the Mahalanobis distance $\| D(\uu_i, \hmay_{ab}) \|_{\Sigma_{ab}} = (\uu_i^a - \uu_i^{a'}) \Sigma^{-1}_{ab} (\uu_i^a - \uu_i^{a'})^T$.  $\Sigma_{ab}$ represents the propagated covariance matrix through the mapping induced by the corresponding homography.} The $\nabla_{\mathcal{N}}$ operator takes in to account the variation over the label function on a specified neighbourhood $\mathcal{N}$. In this paper we use a 4-connectivity neighbourhood although it can be easily extended to any connectivity pattern. $\omega_{\mathcal{N}}$ penalises pixels that are far away in terms of Euclidean distance. Model initialisation is carried out by applying the DLT algorithm \cite{hartley2003multiple}.

We analyse the accuracy of CORAL by conducting a simulated experiment with perfect data associations between views and perfect labelling. Further, we show the performance of our approach on the extensively validated Adelaide benchmark dataset \cite{wong2011dynamic}.    

\subsubsection{Simulation Experiments}

To characterise the performance and robustness of CORAL with respect to existing algorithms, experiments were first run in a controlled simulation environment. The simulation environment consists of three planes, placed mutually orthogonally to each other. This configuration resembles, for instance, the end of a corridor or corner of a building whereby two walls and the ground are simultaneously observed. Uniform sampling of the planes creates the point features observable from two frames by a camera with associated noise $\sigma_{pixel}$. Samples of the simulation environment are shown in Table \ref{fig:noisequality}.

In addition to points directly sampled from planes, we add outliers at different percentages to the simulation. To do so, an uniform sampling of the scene space is performed. Then, points are similarly projected to the image planes. While these outlier points naturally account for points that do not originate from a plane, we argue that they also include points that are wrongly matched between image frames. A triangulation of a wrong match between frames leads to a point that lies arbitrarily in space even if one or both of its corresponding image points lied on a plane. This corresponds to our outlier generation model.

We first tested CORAL under different values of $\sigma_{pixel}$ in the absence of any outliers. This was compared to a implementation of sequential RANSAC. In addition, we compare to PEARL using the open-source implementation available in \cite{pearlcode}. The Misclassification Error (ME) is used as metric to quantify the accuracy.
\begin{equation}
    \mathbf{ME}=\frac{\# \textrm{Misclassified points}}{\#\textrm{points}}
    \label{eq:merror}
\end{equation}
\begin{figure}[t]
\begin{center}
   \includegraphics[width=\linewidth]{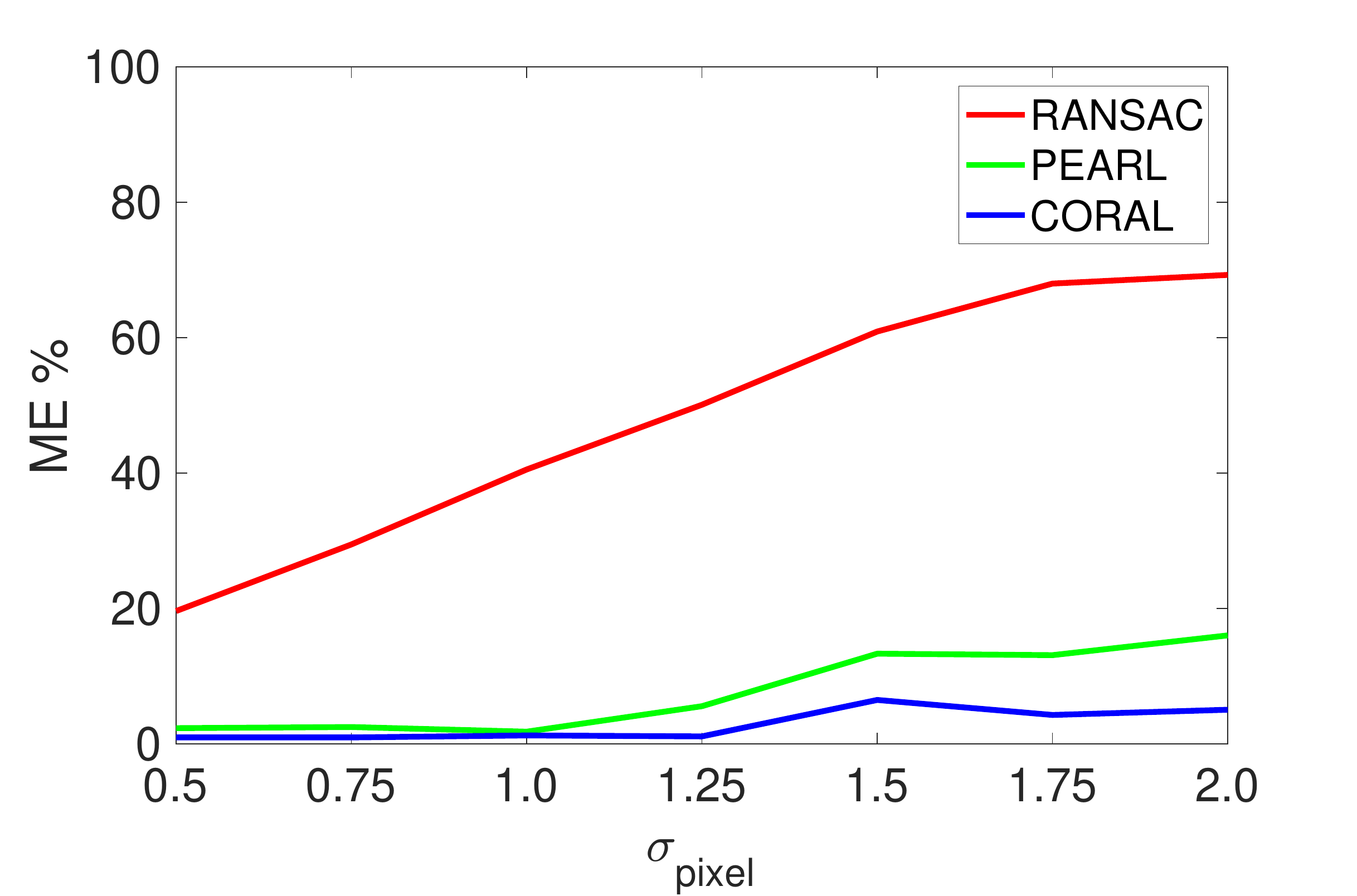}
\end{center}
   \caption{Misclassification error against sensor noise $\sigma_{pixel}$ in the absence of outliers. The errors are obtained for RANSAC , PEARL and CORAL algorithms. The ME is evaluated and averaged over ten different two-view pairs of the simulated environment.}
\label{fig:simnoise}
\end{figure}
\begin{figure}[t]
\begin{center}
   \includegraphics[width=0.95\linewidth]{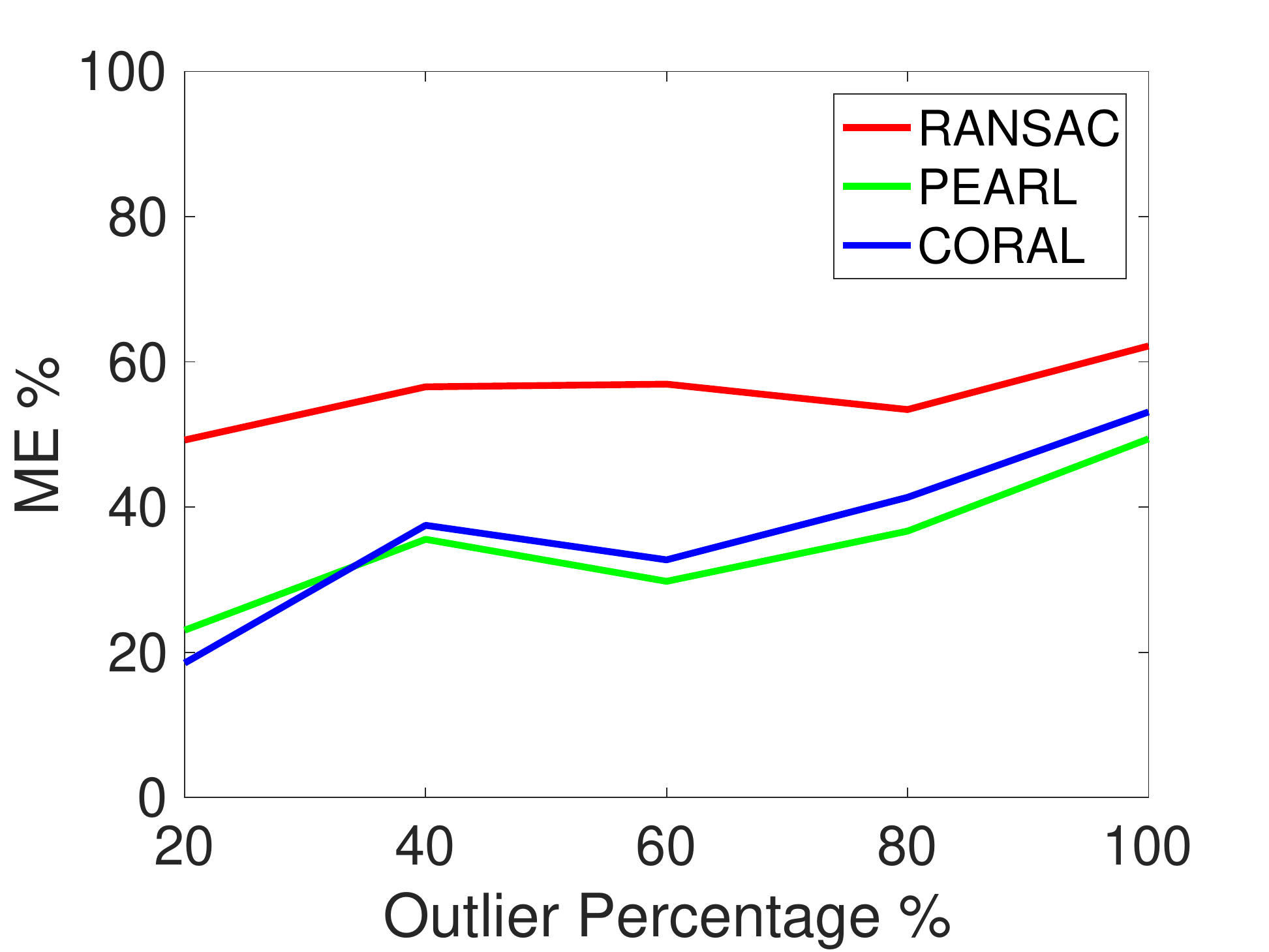}
\end{center}
   \caption{Mislabelling error against outlier percentage for the three methods. Evaluated  and averaged over ten different views.}
\label{fig:simoutliers}
\end{figure}

Results from the three approaches are shown in Figure \ref{fig:simnoise}. It is clear that the global energy approaches outperform RANSAC as the sensor noise increases, in fact CORAL reports better results than PEARL for the largest noise value. The higher ME reported by RANSAC in the presence of noise can be explained by analysing Table \ref{fig:noisequality}. The table shows the triangulated noisy points in the simulation environment for $\sigma_{pixel}=[0.5,1.5,1.5]$ together with the ground truth planes from which they were originally sampled from. Colour coding of membership to a particular model is used to show the results of the multi-homography extraction for the three stated approaches.

\begin{table*}[ht]
    \centering
    \begin{tabular}{ c c c}
        $\sigma_{pixel}=0.5$ & $\sigma_{pixel}=1.0$ & $\sigma_{pixel}=1.5$ \\ 
        
    \includegraphics[width=0.3\textwidth]{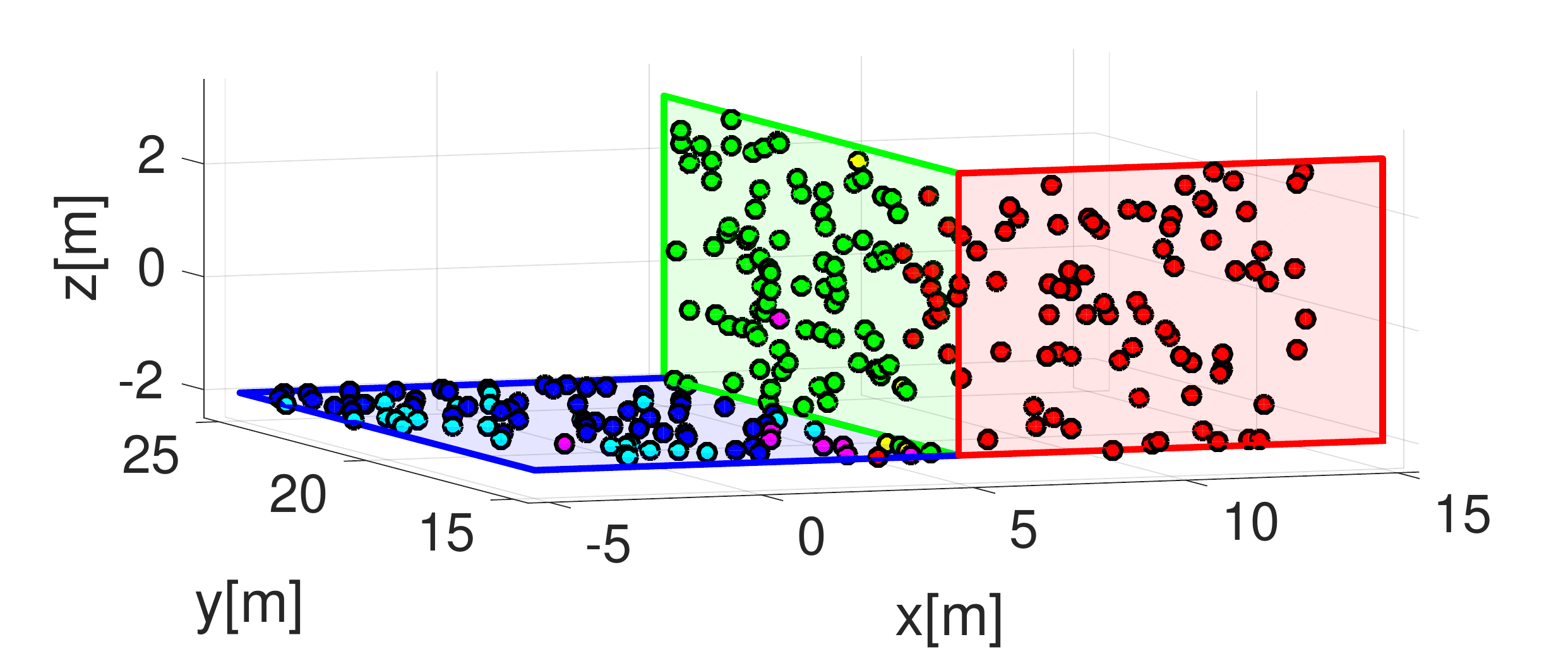} &
    \includegraphics[width=0.3\textwidth]{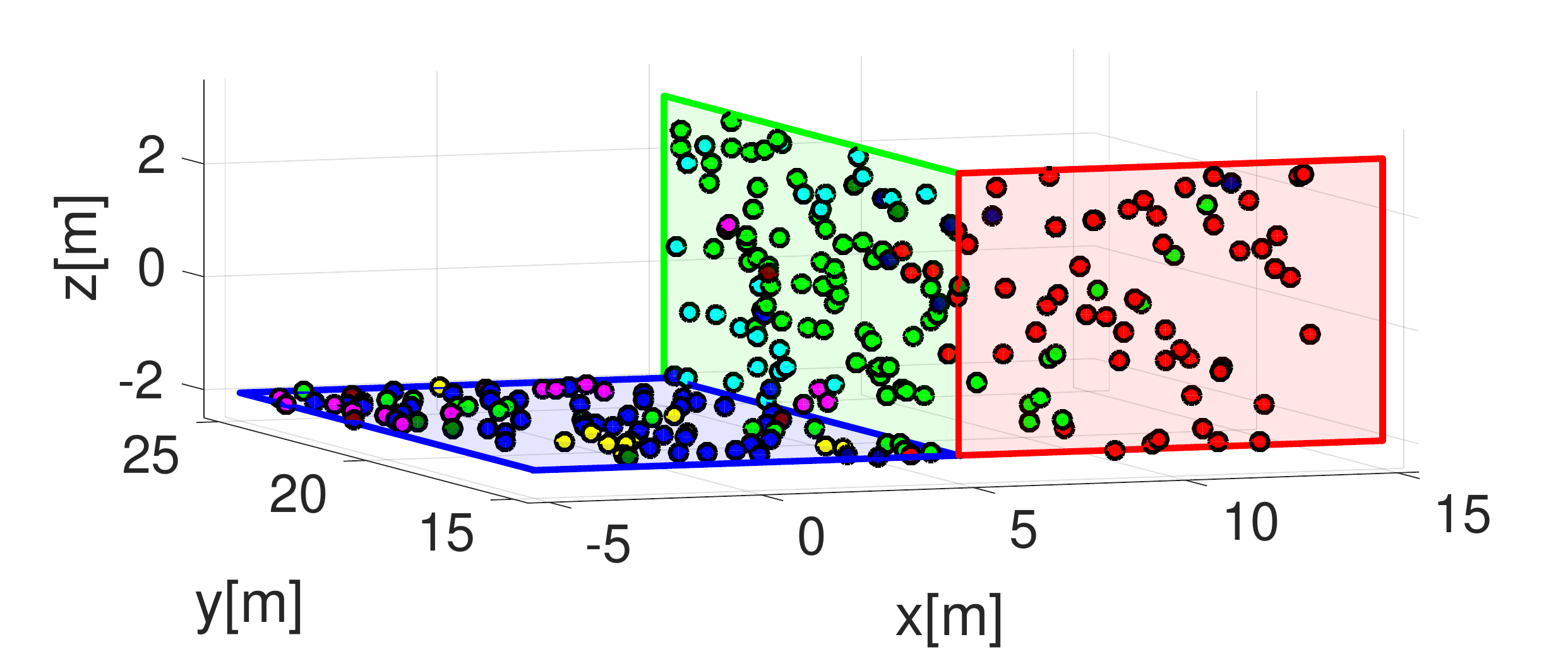} &
    \includegraphics[width=0.3\textwidth]{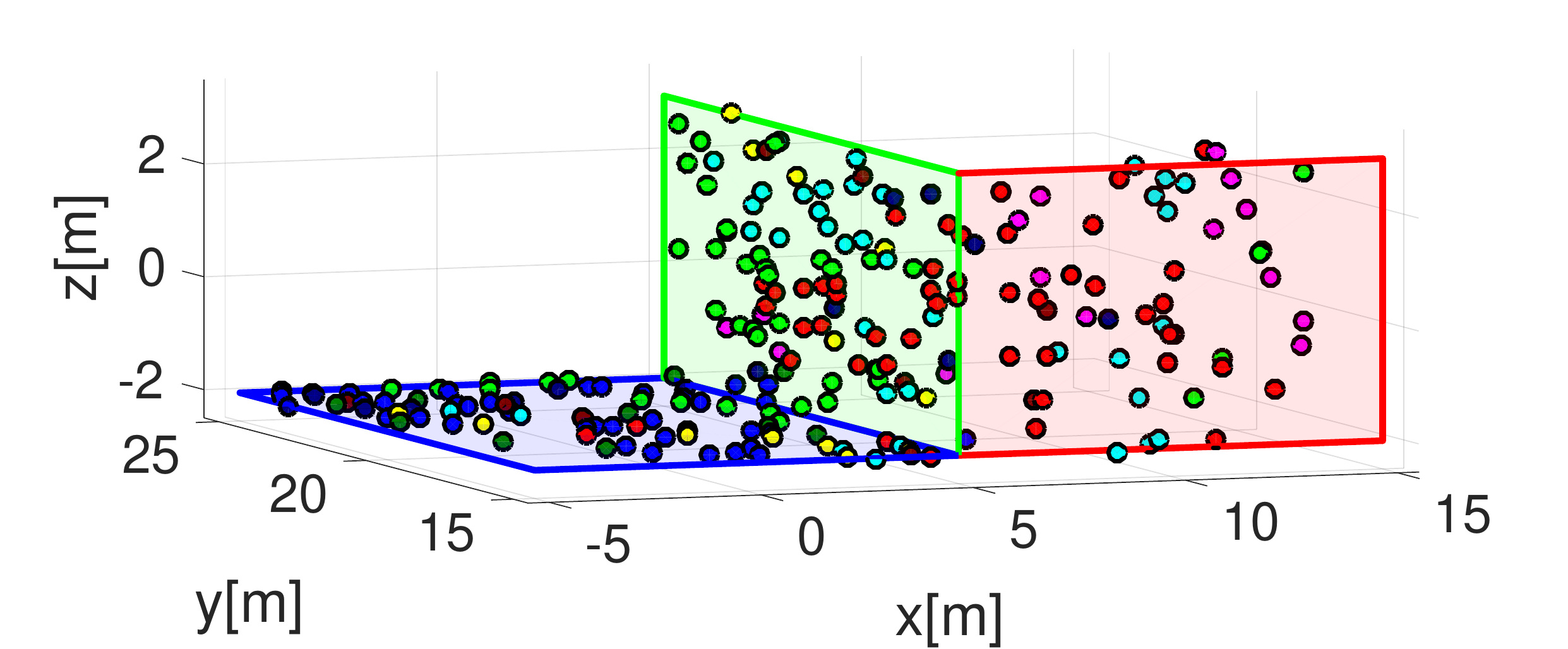} \\

    \includegraphics[width=0.3\textwidth]{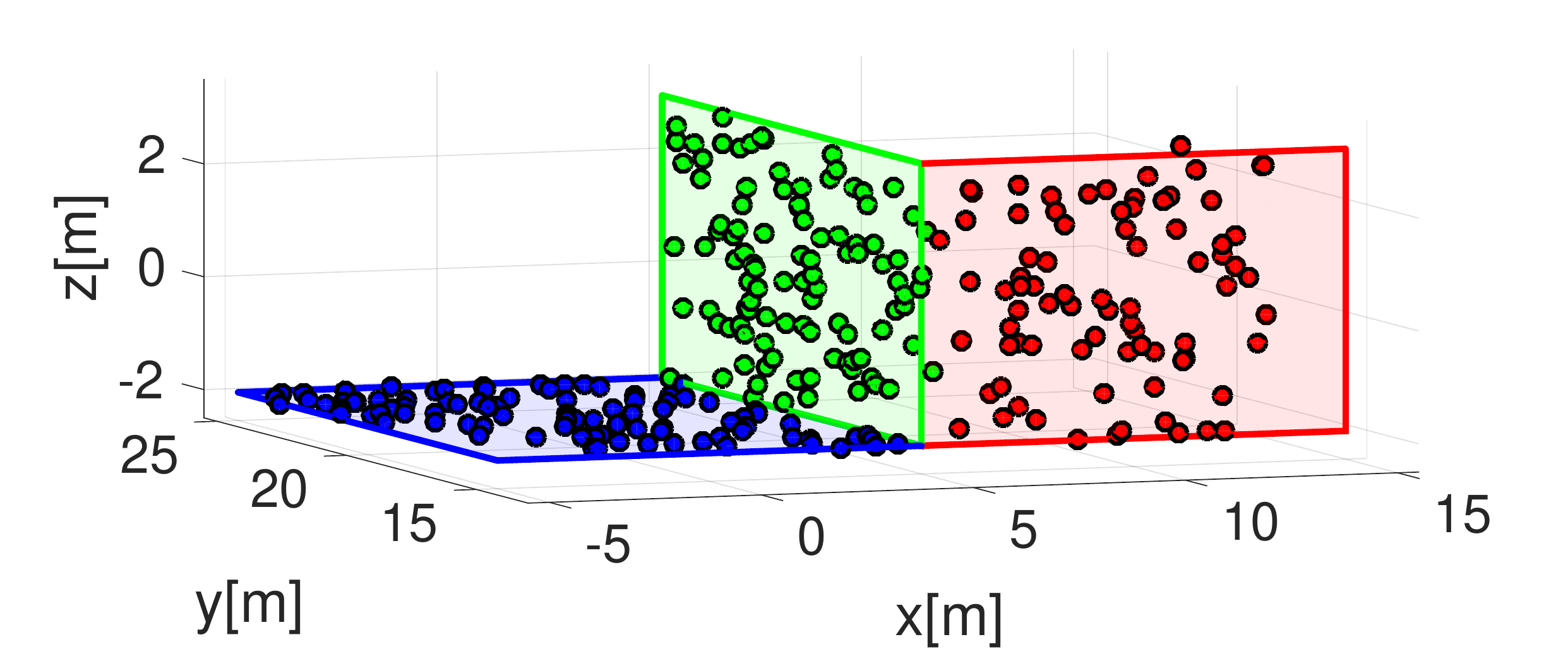} &
    \includegraphics[width=0.3\textwidth]{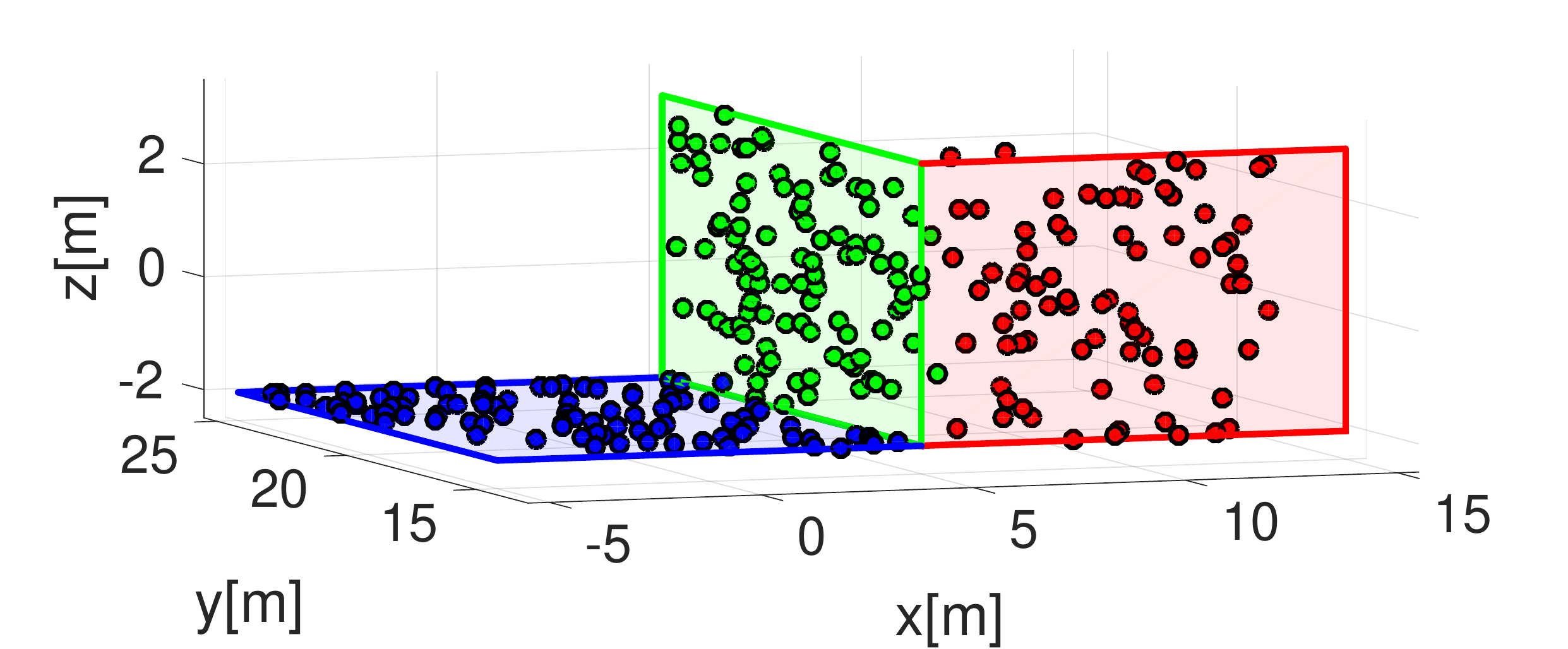} &
    \includegraphics[width=0.3\textwidth]{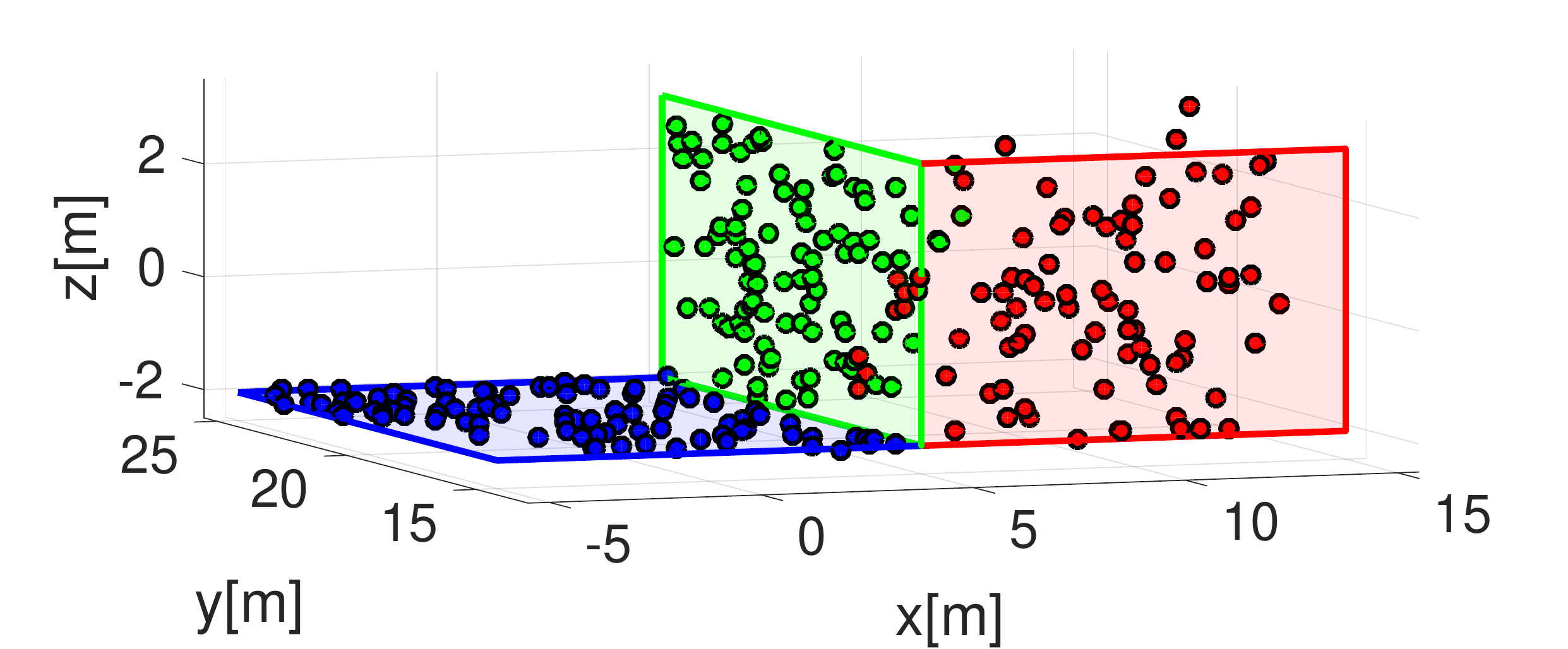} \\

    \includegraphics[width=0.3\textwidth]{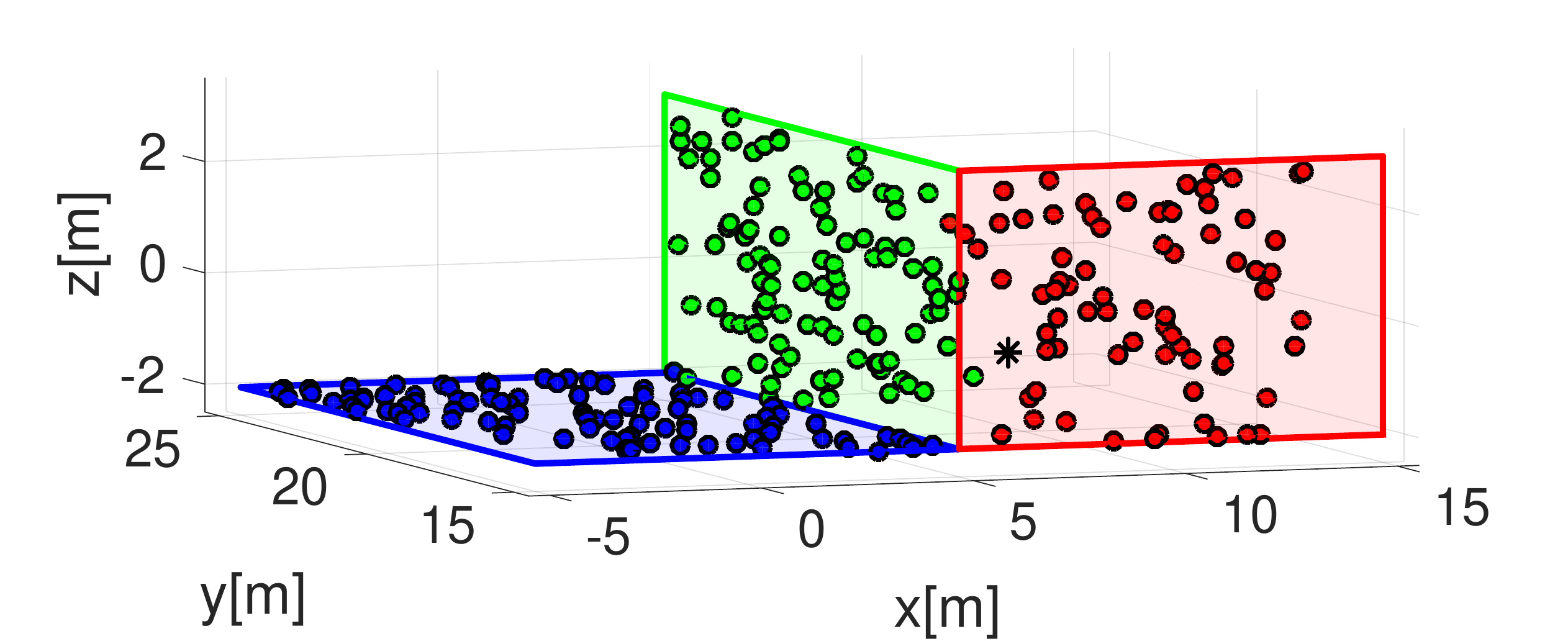} &
    \includegraphics[width=0.3\textwidth]{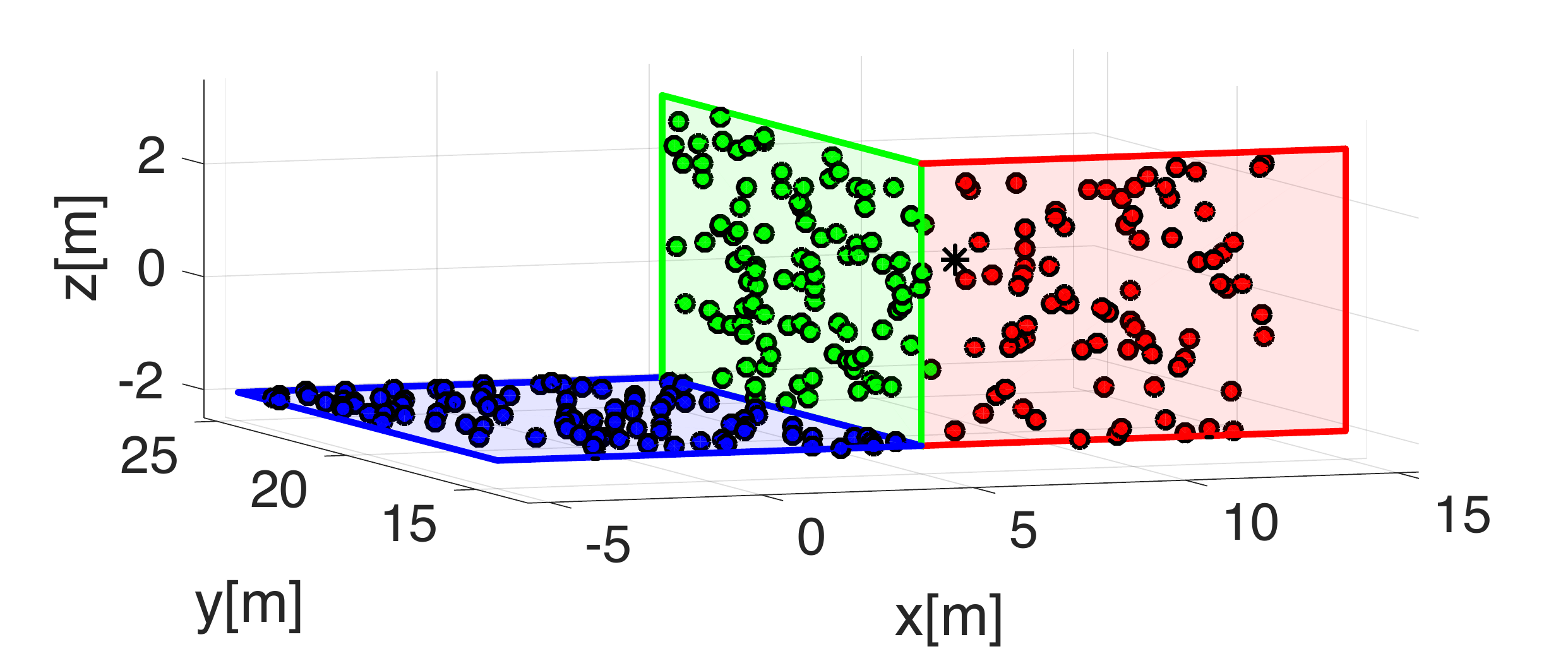}  &
    \includegraphics[width=0.3\textwidth]{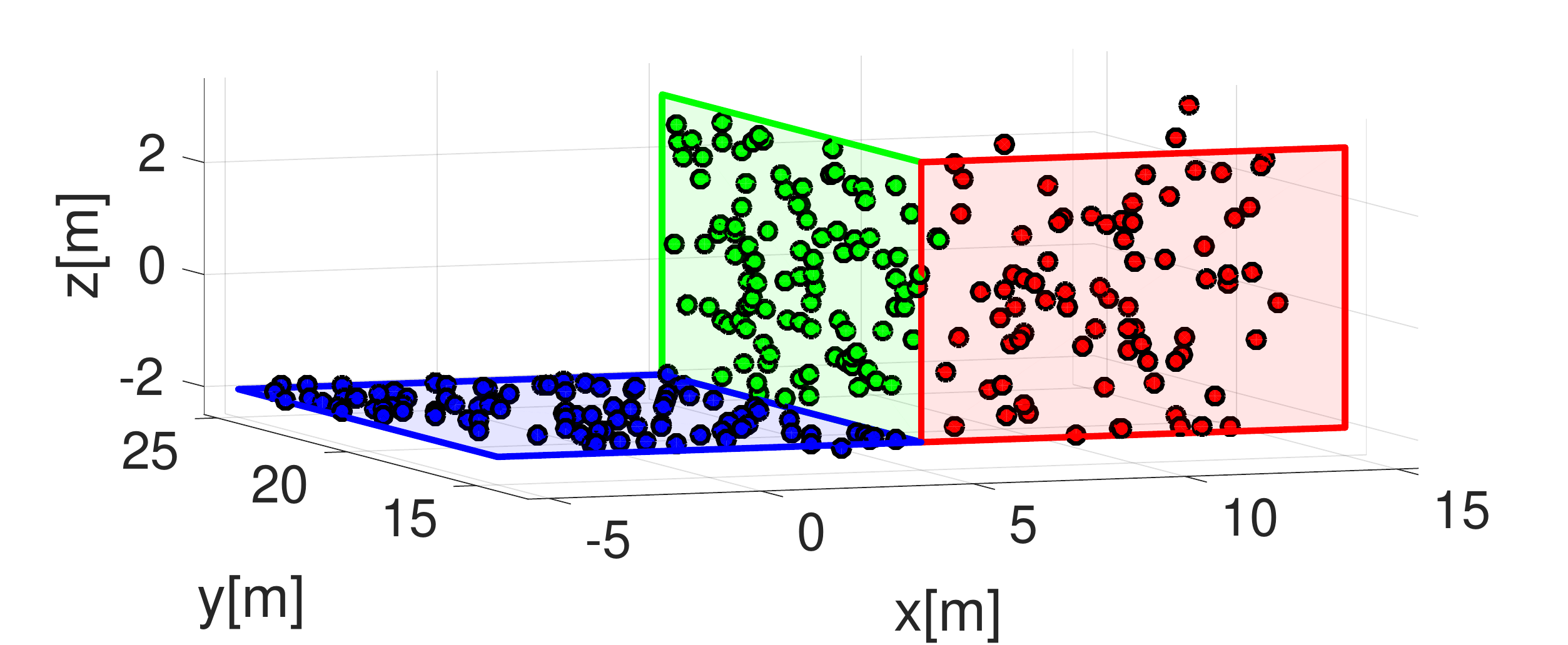} \\

    \end{tabular}
    \caption{ The triangulated positions of noisy pixels together with the ground truth planes (red, green,and blue patches) are shown under different values of $\sigma_{pixel}$. The results of the multi-homography detection are displayed by a colour coding the point classification to their assigned models. RANSAC, PEARL and CORAL results are shown in the first, second and third row respectively.}
    \label{fig:noisequality}
\end{table*}
For $\sigma_{pixel}=1.5$ RANSAC selects models that, though geometrically valid, are not consistent with the ground truth planes. See for example the combination of green and red points on the two vertical planes. The energy approaches are more robust to this kind of noise as the smoothness prior ensures convergence to a better solution consistent with the ground truth. This is shown in the last column of Table \ref{fig:noisequality}. 

We also test our proposed formulation in terms of robustness against outlier contamination. In this work the outlier percentage is defined as the ratio of outliers to inliers. The results of the three methods are summarised in Figure \ref{fig:simoutliers} with the sensor noise set to $\sigma_{pixel}=1.0$. For low percentage values $< 60\%$ the energy methods are more robust than RANSAC, with PEARL exhibiting a small advantage over CORAL. At higher percentages, it becomes much more difficult to obtain the solution and the ME for all the different approaches appear to converge. This is due to an inherent bias in the ME given that the outlier model in this case, dominates the data. Although RANSAC can incorrectly label models, it would still reject outliers resulting in a lower ME as the outlier class increases.

\subsubsection{Experiments with real data} 
Following the simulation verfication, we benchmarked our result against the state-of-the-art on real data. The AdelaideRmf dataset \cite{wong2011dynamic} is used in this evaluation. It consists of image pairs with extracted keypoints and manually labelled ground truth. The performance on this dataset of different multiple model fitting approaches is available in \cite{magri2016multiple}. From these results, we carried out a comparison of CORAL to T-linkage \cite{magri2014t}, J-linkage \cite{toldo2008robust}, SA-RCM \cite{pham2014random}, RPA \cite{magri2015robust},  Grdy-RansaCov and ILP-RansaCov \cite{magri2016multiple}. These results are shown in Table \ref{tab:BenchmarkME}.

From Table \ref{tab:BenchmarkME} it can be seen that the best mean performance is achieved using our algorithm, with few cases sharing the best performance.
The better median performance by Multi-H \cite{barath28multi} can be explained by its specialised initialisation using affine transformations that have been shown to be superior to the DLT. This approach then subsequently uses a PEARL formulation for energy minimisation. Our approach would also benefit from this specialised initialisation; however, for better comparison with the state of the art we utilise an initialisation based on the DLT for model generation.

\begin{figure*}
    \centering
    \begin{subfigure}[b]{0.3\textwidth}
        \includegraphics[width=\textwidth]{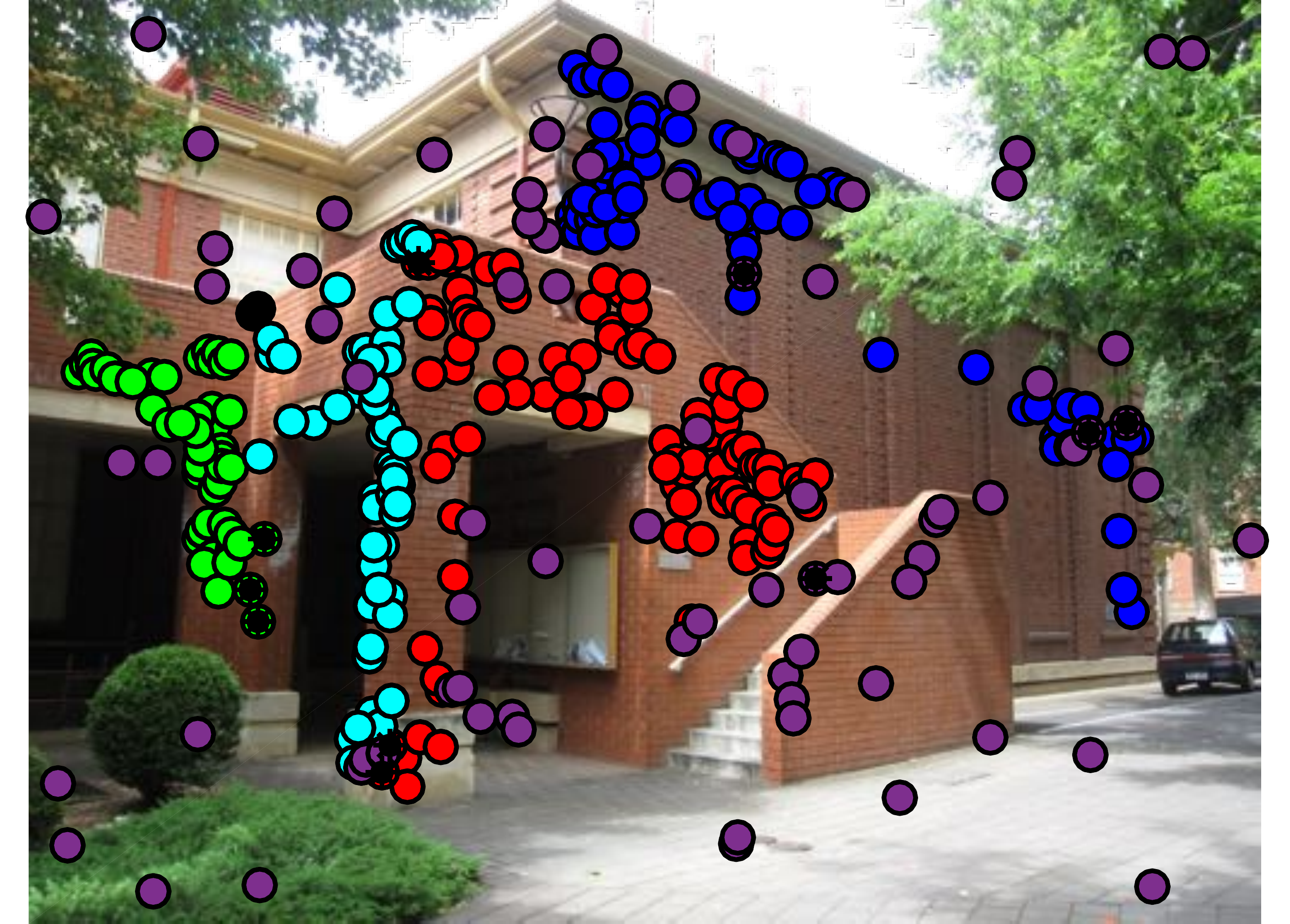}
        \subcaption{Johnsona}
    \end{subfigure}
    ~ 
    \begin{subfigure}[b]{0.3\textwidth}
        \includegraphics[width=\textwidth]{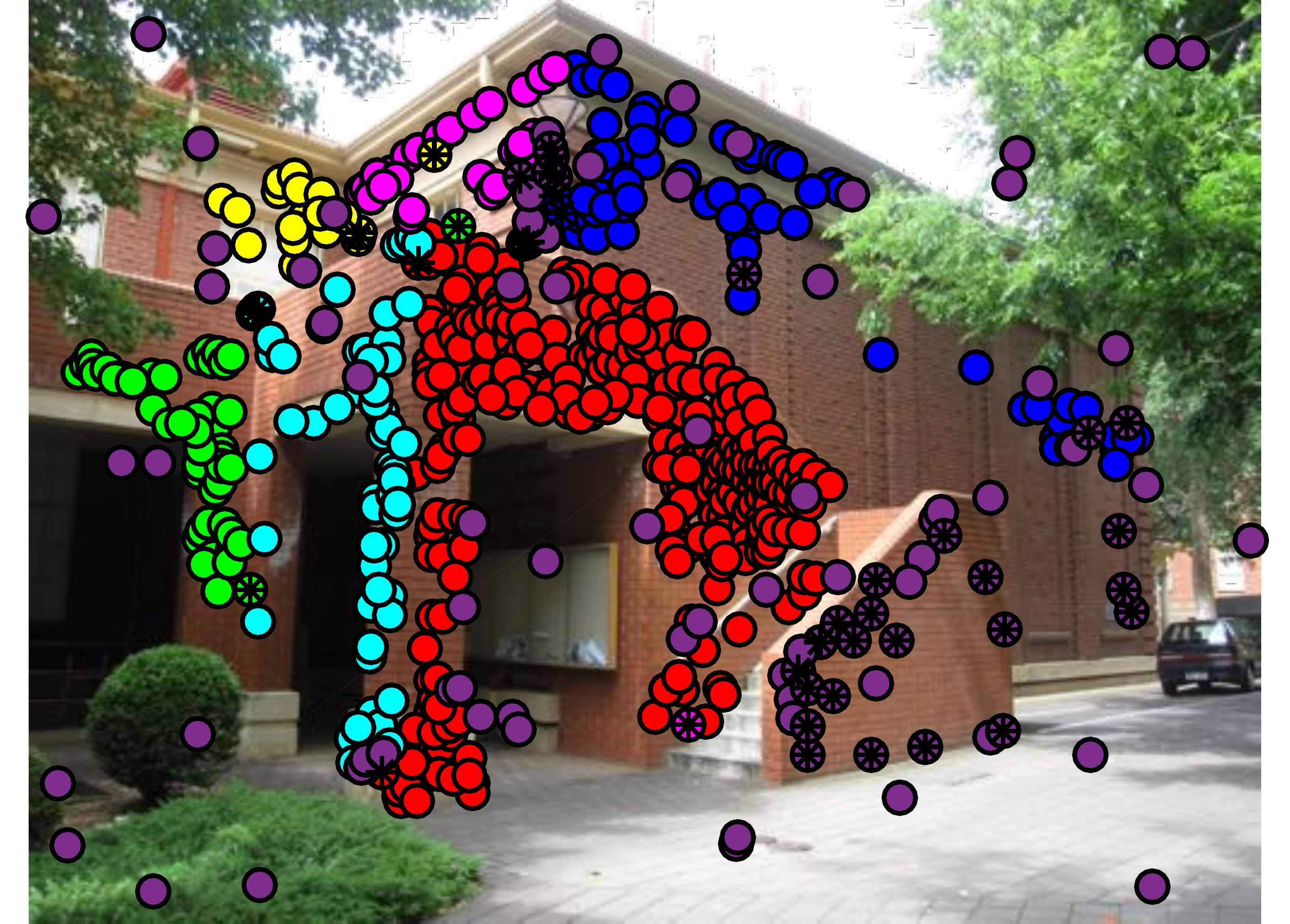}
        \subcaption{Johnsonb}
    \end{subfigure}
    \begin{subfigure}[b]{0.3\textwidth}
        \includegraphics[width=\textwidth]{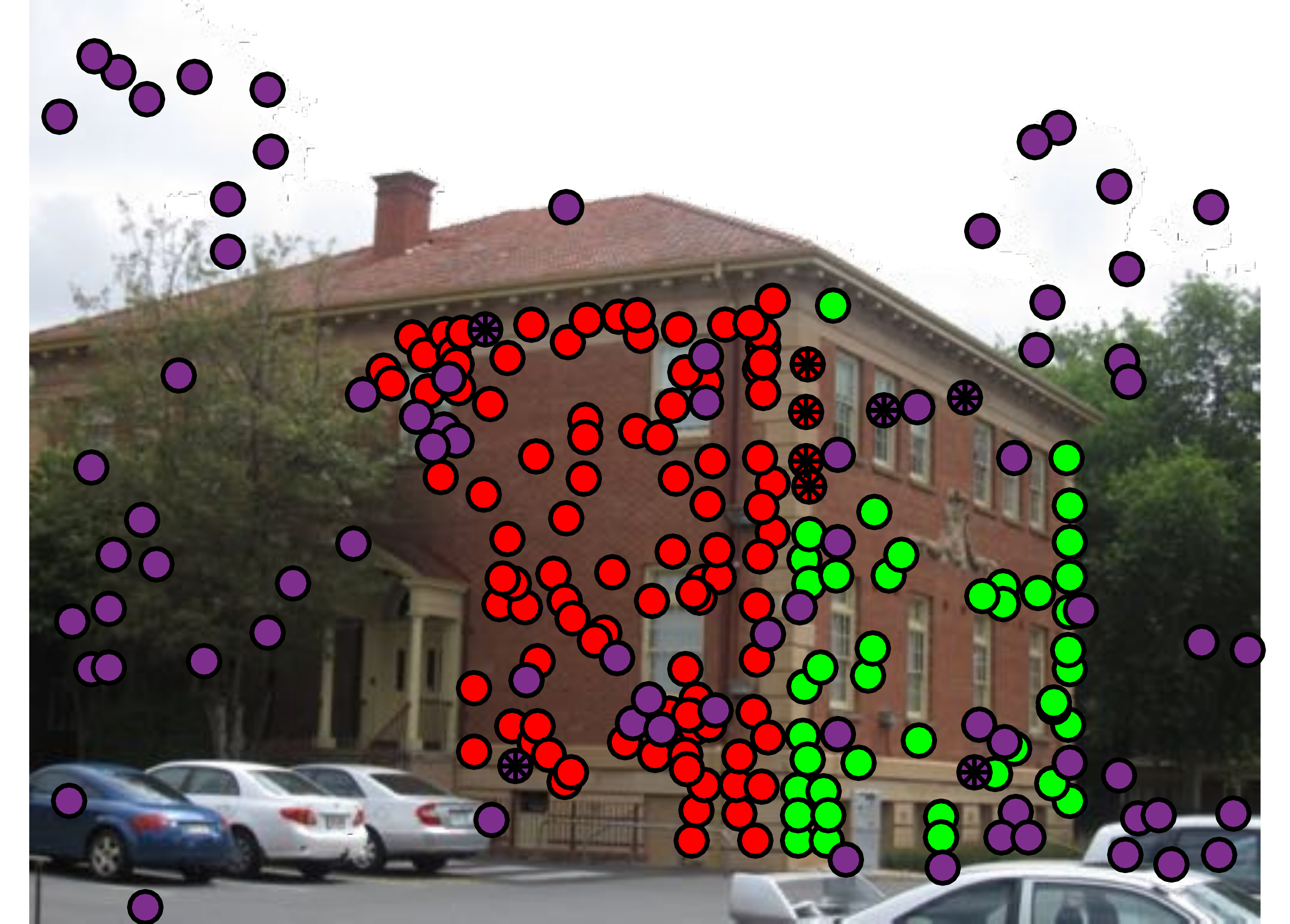}
        \subcaption{ladysymon}
    \end{subfigure}
    
    \caption{Sample of images from the AdelaideRmf Dataset. With the membership to the different homographies indicated by the different colours. Crosses are used to signify points that are mislablled while the uniform outlier label $\emptyset$ is shown in purple through all figures.} \label{fig:benchmarkquality}
\end{figure*}

Some of the images with the detected homographies are shown in Figure \ref{fig:benchmarkquality}. From this Figure it can be seen that our approach is able to deal with a varied range of models that make up the final solution. 


\begin{table*}[t]
    \centering
    \begin{tabular}{ |c|| c| c| c| c| c| c| c| c|}
    \hline
          & J-Lnkg & T-Lnkg & RPA & SA-RCM & Grdy-RansaCov & ILP-RansaCov & Multi-H & \textbf{CORAL} \\ \hline
        mean    & 25.50 & 24.66 & 17.20 & 28.30 & 26.85 & 12.91 & 4.40 & \textbf{4.2117}  \\ \hline
        median   & 24.48 & 24.53 & 17.78 & 29.40 & 28.77 & 12.34 & \textbf{2.41} & 3.48  \\ \hline
    \end{tabular}
    \caption{Misclassification error for two view plane segmentation in the AdelaideRmf dataset. Part of the results in this table are reported in \cite{magri2016multiple}. }
    \label{tab:BenchmarkME}
\end{table*}

\subsection{Plane detection with RGB-D images}

The second case of structure detection is the extraction of planes from a single RGB-D image. In this scenario, we can exploit the regularity of the image pixel grid. This idea, in contrast to a 3D point cloud representation, will fully leverage the effects of the smoothness regularisation term. In other words, the individual pixel solutions will propagate along the domain of the image with a more clear definition of the local neighbourhood. In order to apply this idea, we chose to work on an inverse depth representation in a per-pixel basis. It can be shown that if two pixels $\uu$ and $\uu^*$ belong to the same planar surface in 3D, their inverse depths $\xi(\uu)$ and $\xi(\uu^*)$ satisfy the following equation
\begin{equation}
    \xi(\uu)-\xi(\mathbf{u}^{*})=\langle \mathbf{w},\mathbf{u}-\mathbf{u}^{*} \rangle
\end{equation}
where $\langle \cdot,\cdot \rangle$ represents the inner product between two vectors. $\mathbf{w} = (w_u,w_v) $ codifies the projection of the 3D plane normals into the image plane. A proof of the validity of this expression is given in the supplementary material. 

In this application, the energy to be minimised is given by
\begin{align} \label{eq:energy_depth}
    &\sum_{l=1}^L \left(\int_{\Omega} (\|\xi(\uu)-\xi(\uu^{*}) - \langle \ww, \uu - \uu^{*} \rangle \|_{\sigma_{\xi}})\phi_i(\uu) d\Omega \right.  \nonumber \\
    & \left. + \lambda \int_{\Omega} \omega_{\mathcal{N}}^\alpha |\nabla_{\mathcal{N}} \phi(\uu) |_{1,2} d\Omega \right) + \beta L 
\end{align}
where $\omega_{\mathcal{N}}^\alpha = e^{-\| \nabla_{\mathcal{N}} I(\uu)\|_{\alpha}}$ and $I$ represents the intensity of corresponding intensity image. These weights serve as a measure of edginess that can be controlled with parameter $\alpha$, thus aiding to preserve sharp discontinuities between objects. Note that we use the Mahalanobis distance on the inverse depth as data term. The rest of the elements in this equation are similar to the general-purpose energy proposed in Equation \ref{eq:energy}.

\begin{table*}[ht]

    \centering
    \begin{tabular}{ c c c c}
    Ground Truth & RANSAC & PEARL & CORAL \\
    \includegraphics[width=0.24\textwidth]{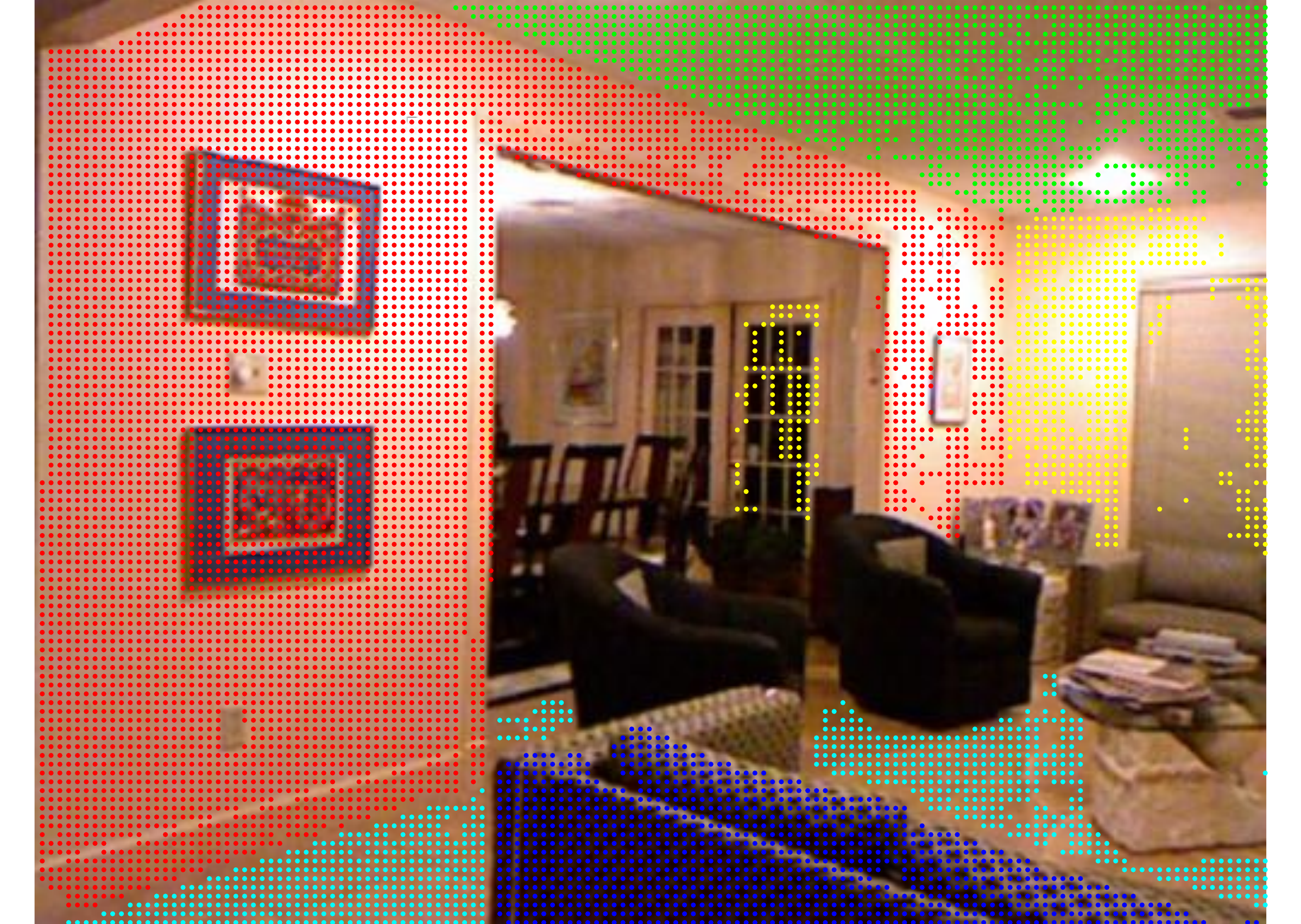} &
    \includegraphics[width=0.24\textwidth]{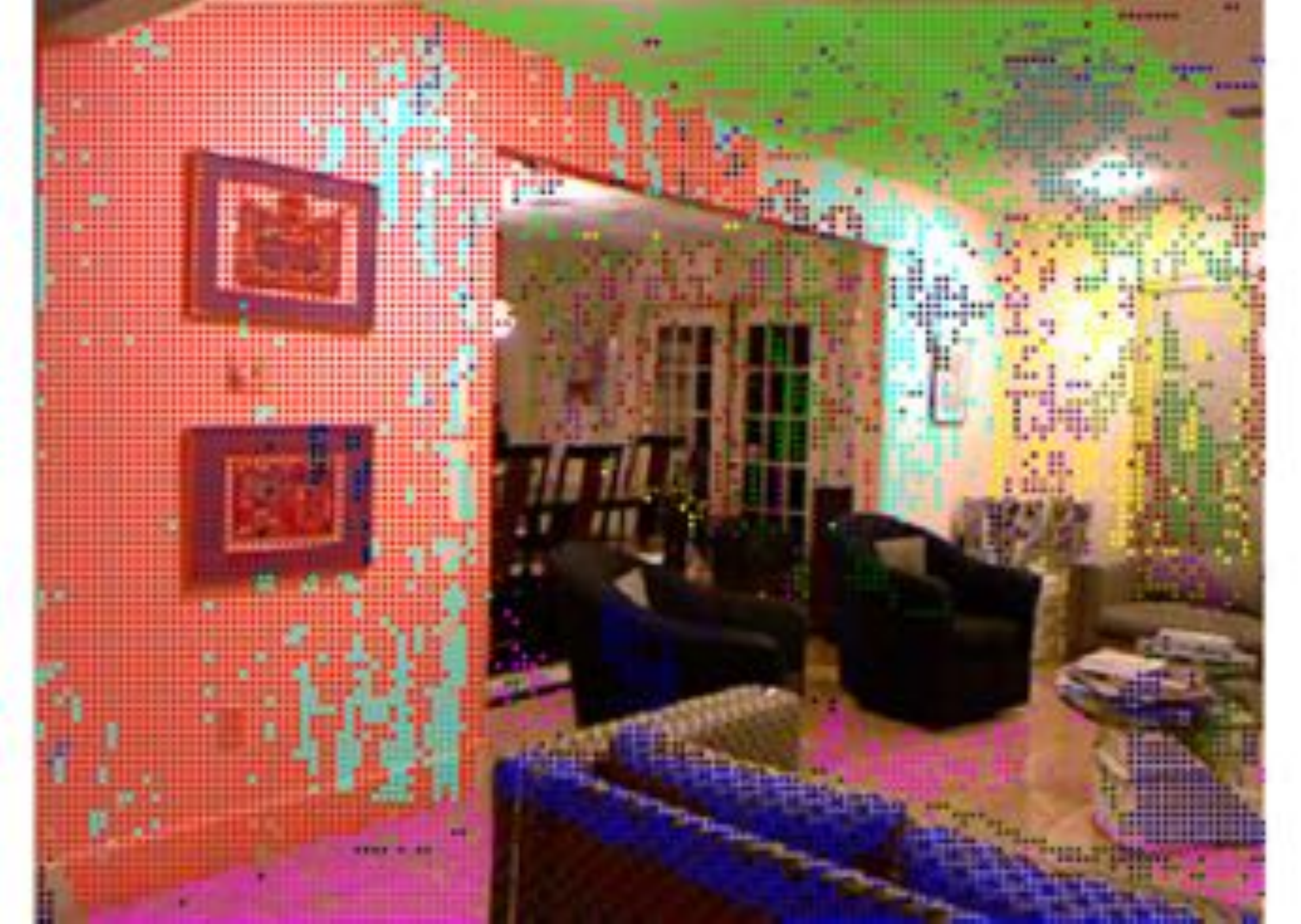} &
    \includegraphics[width=0.24\textwidth]{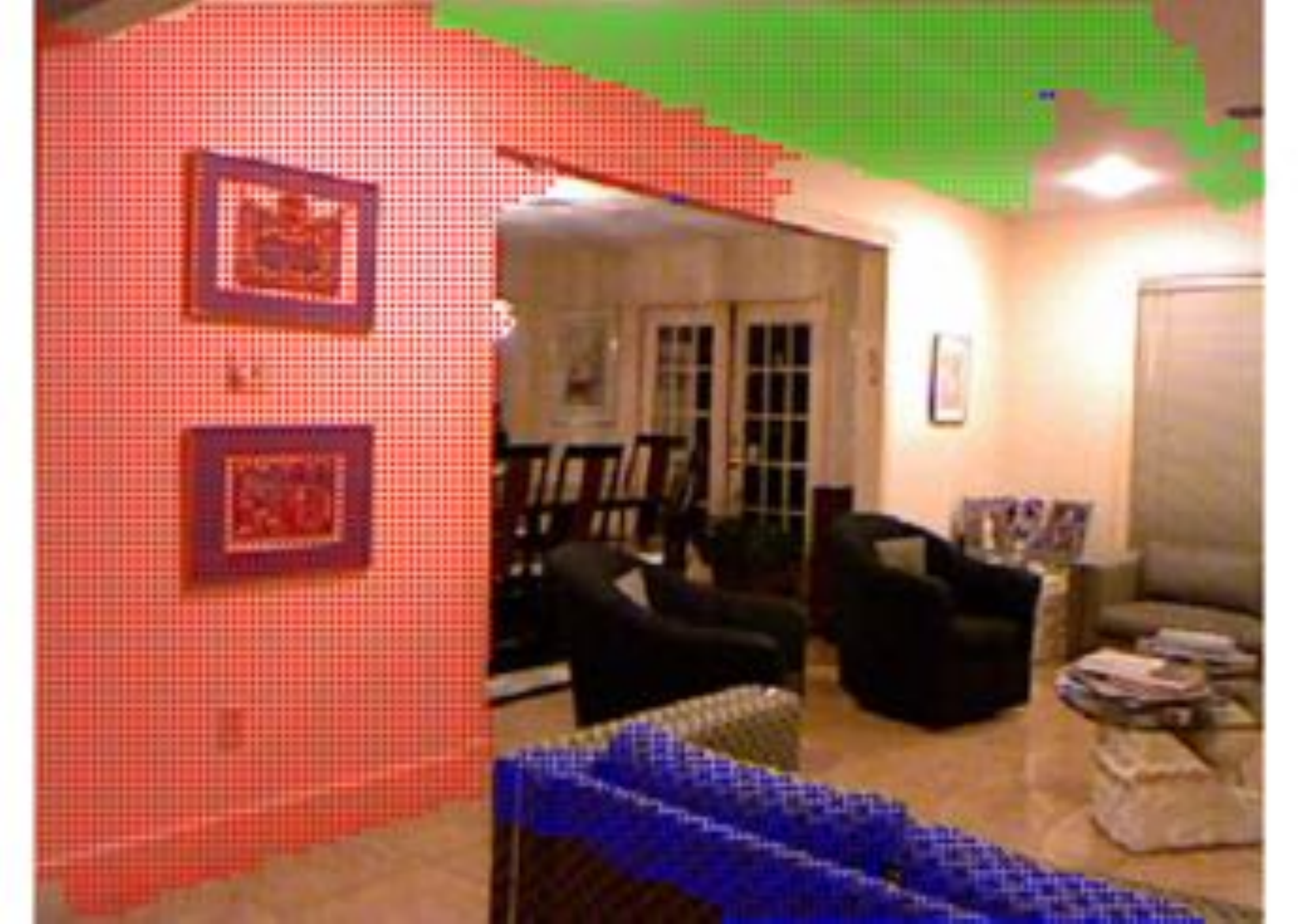} &
    \includegraphics[width=0.24\textwidth]{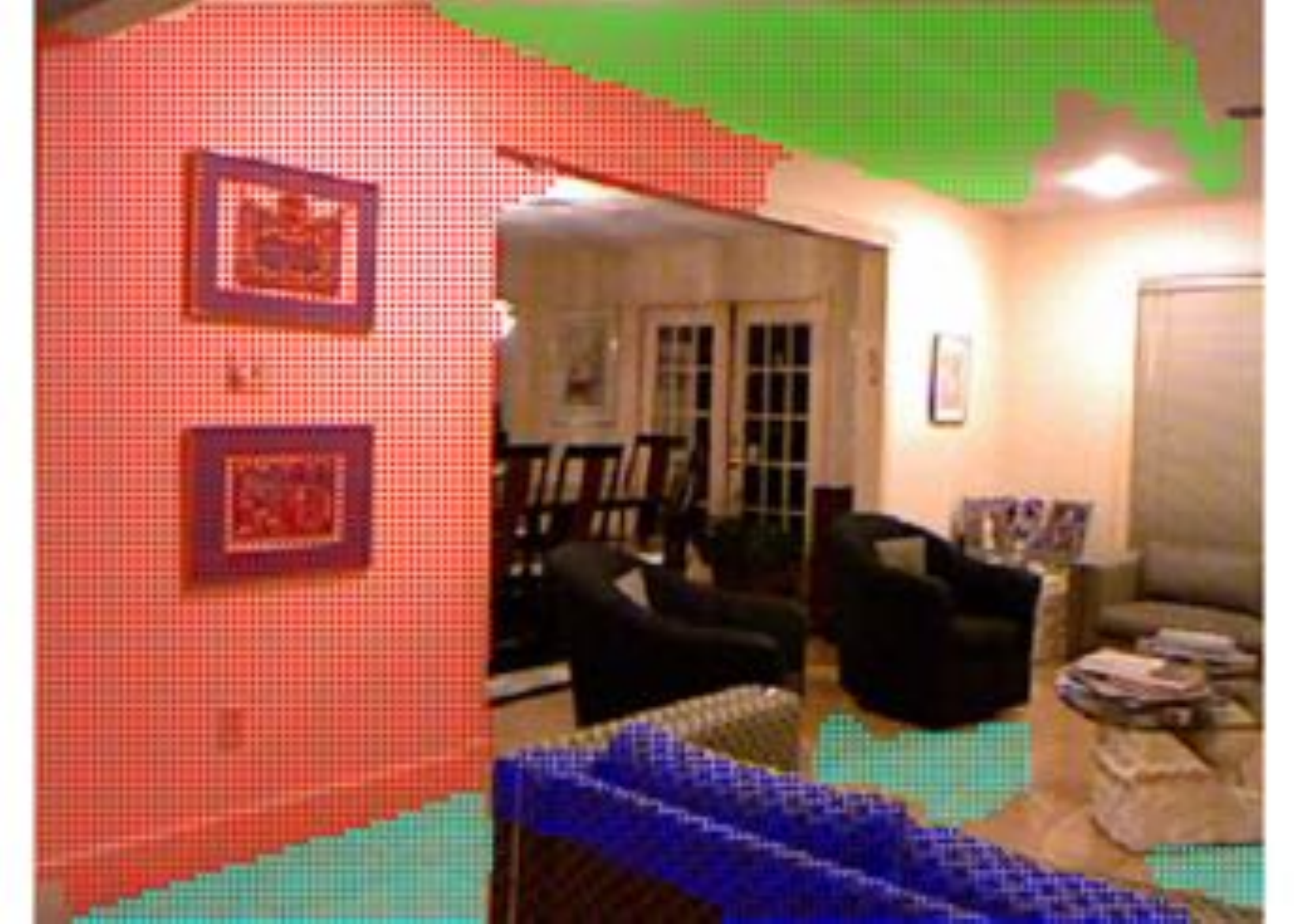} \\
    
    \includegraphics[width=0.24\textwidth]{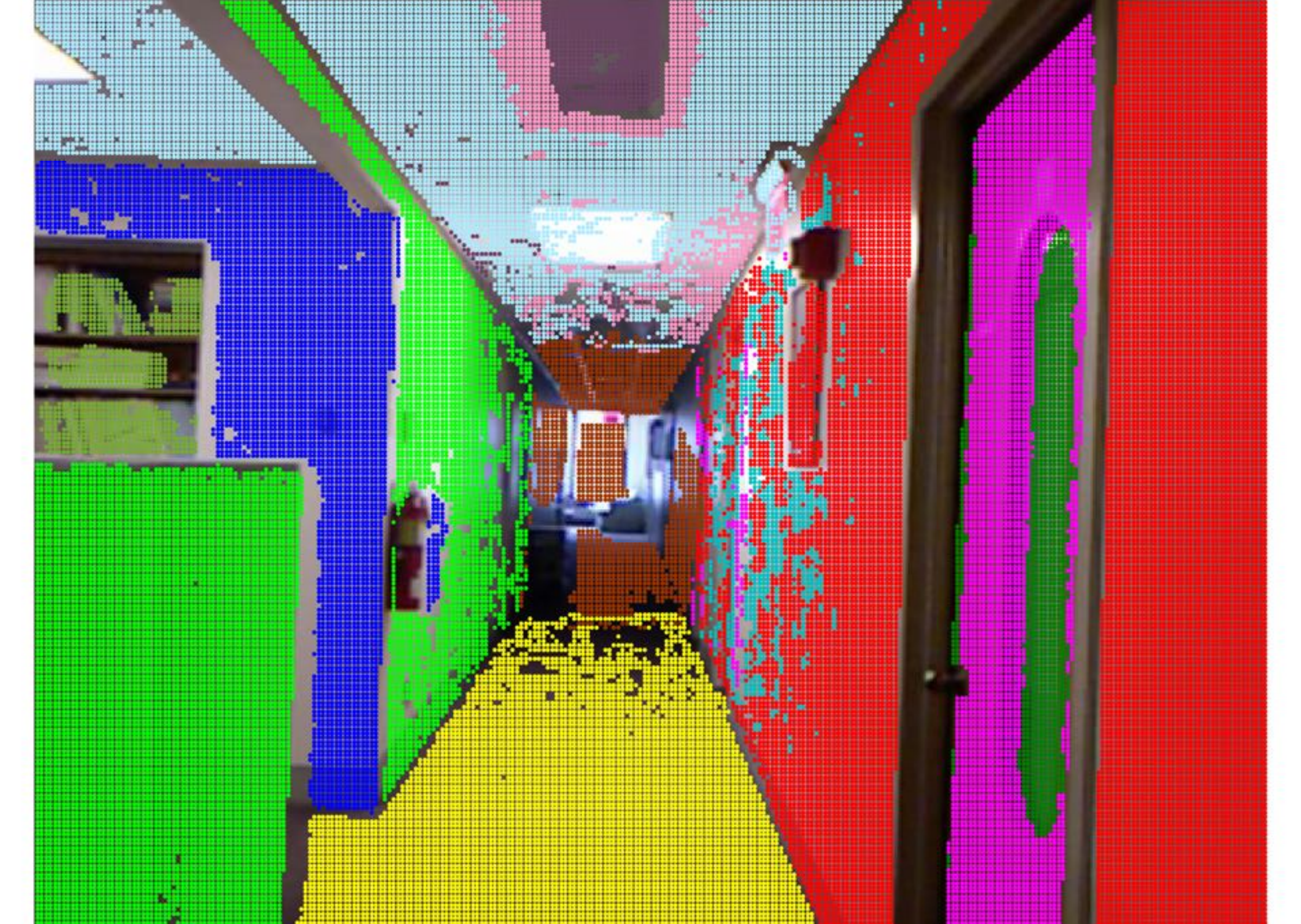} &
    \includegraphics[width=0.24\textwidth]{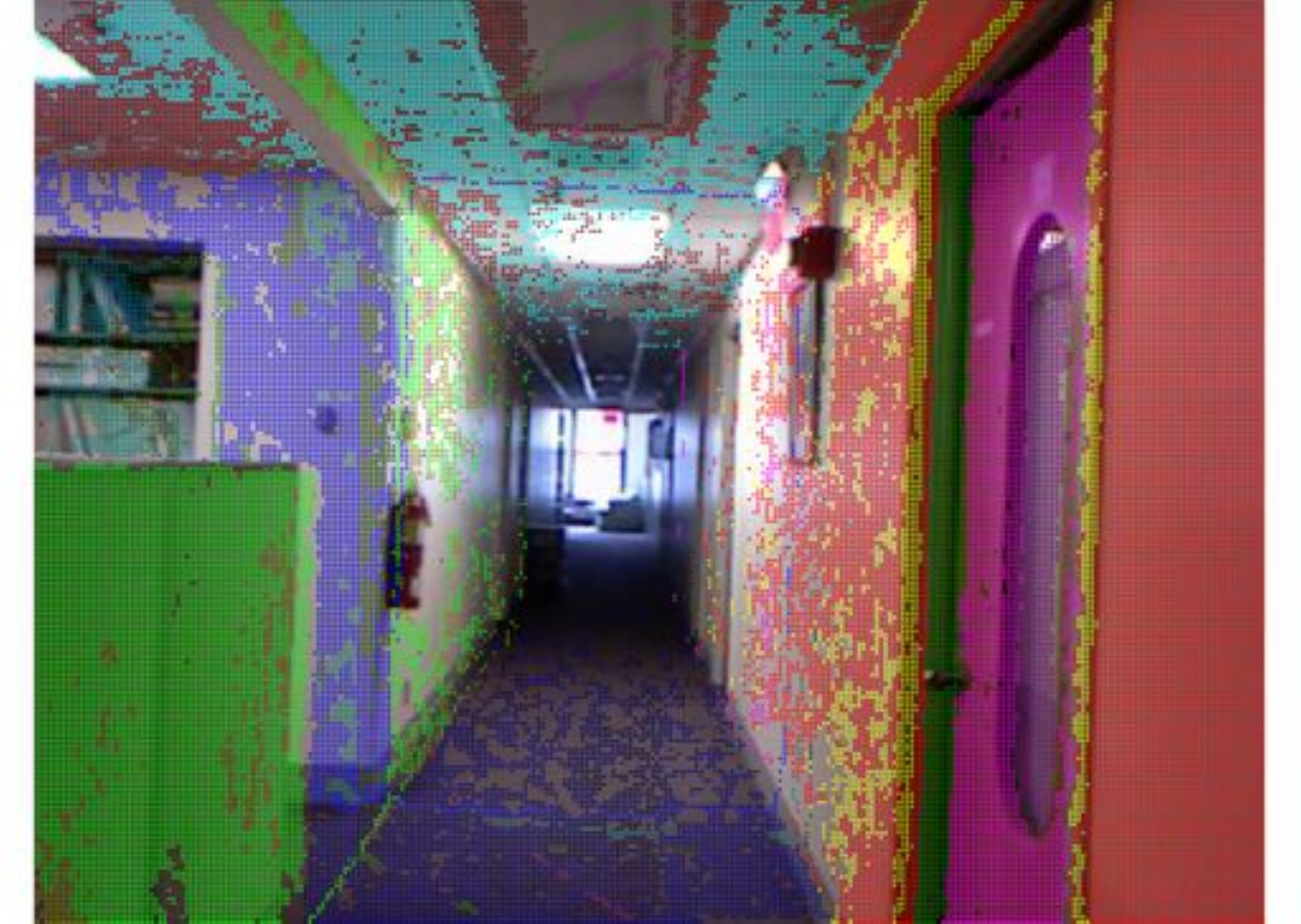} &
    \includegraphics[width=0.24\textwidth]{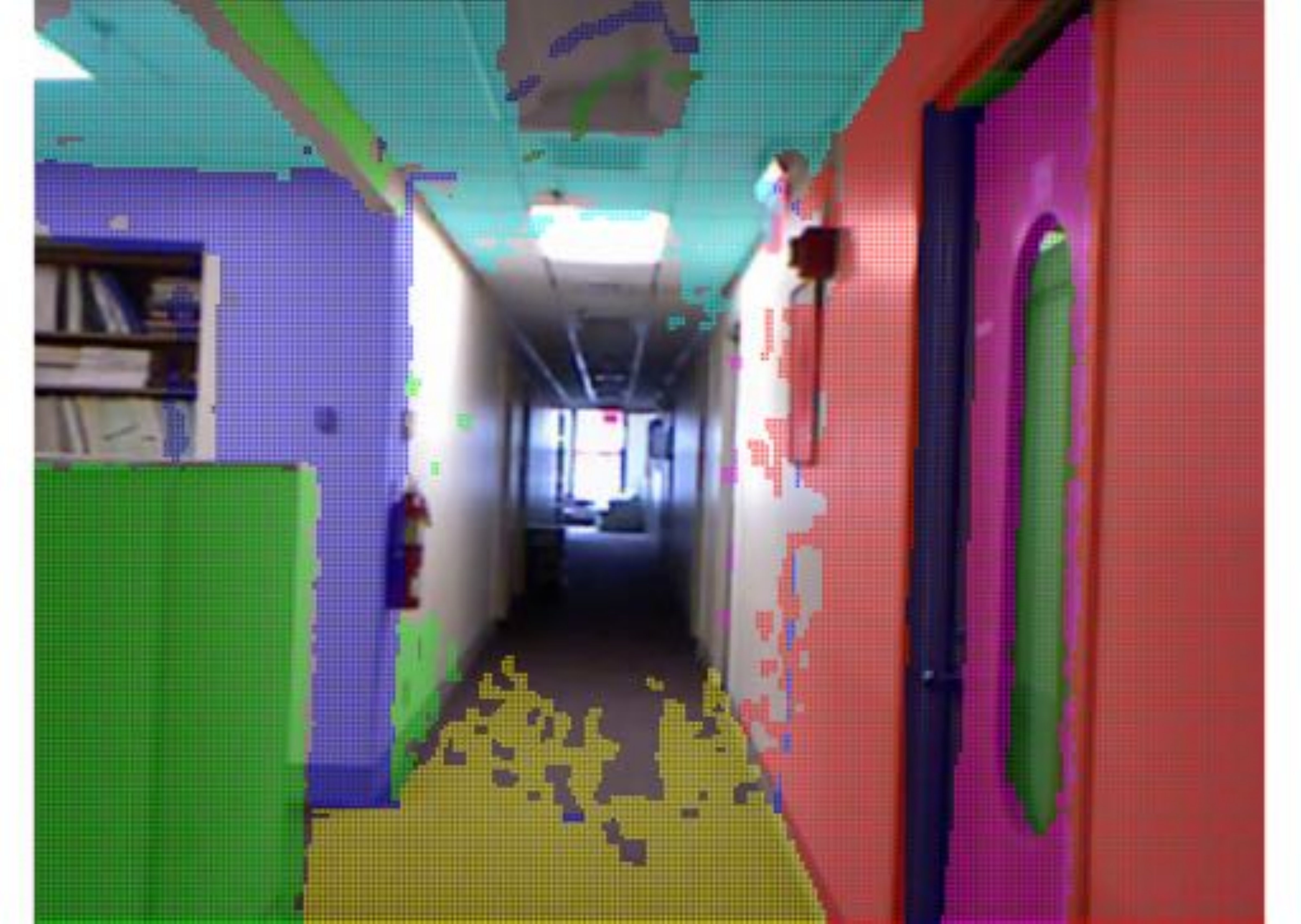} &
    \includegraphics[width=0.24\textwidth]{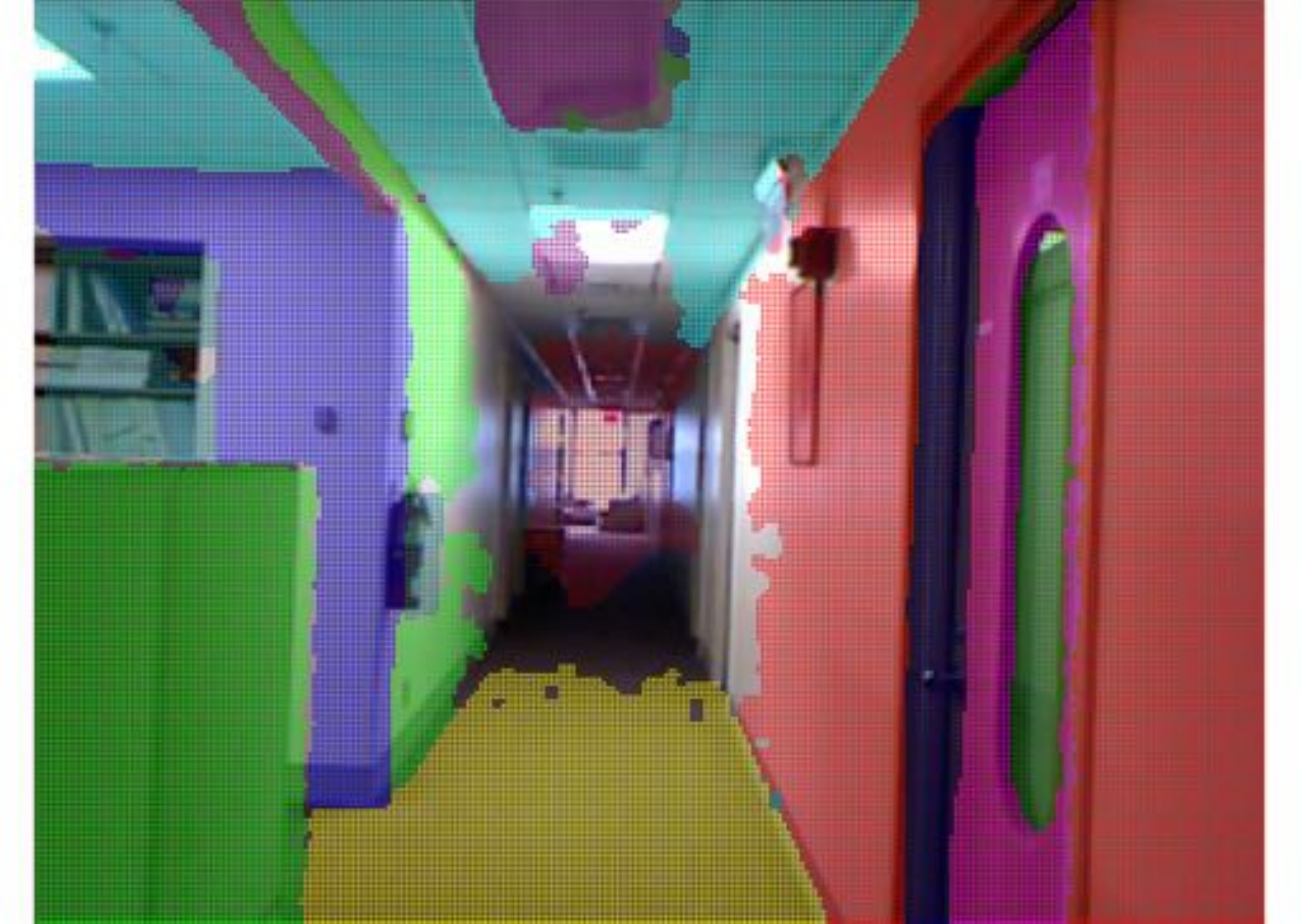} \\
    
    \includegraphics[width=0.24\textwidth]{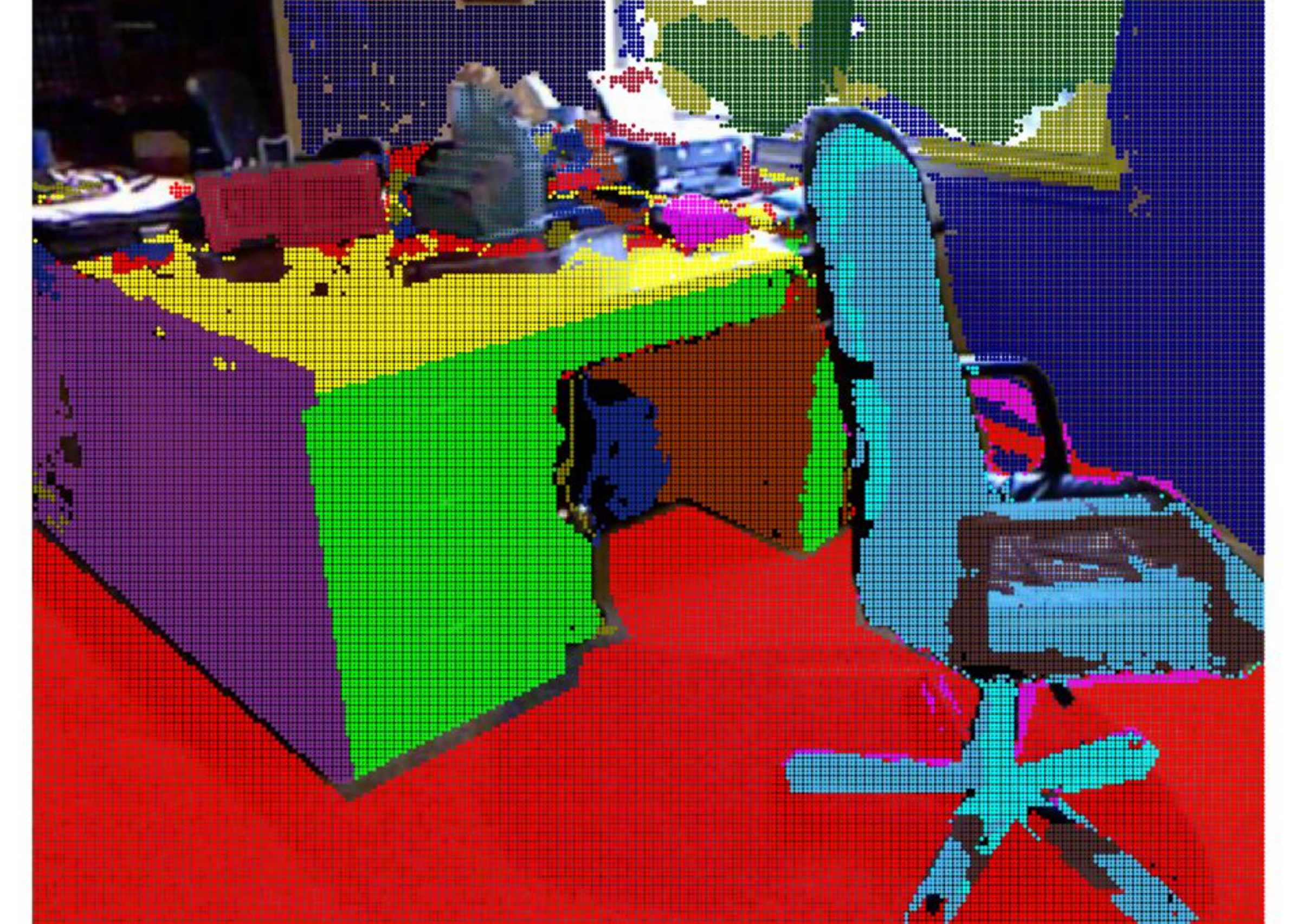} &
    \includegraphics[width=0.24\textwidth]{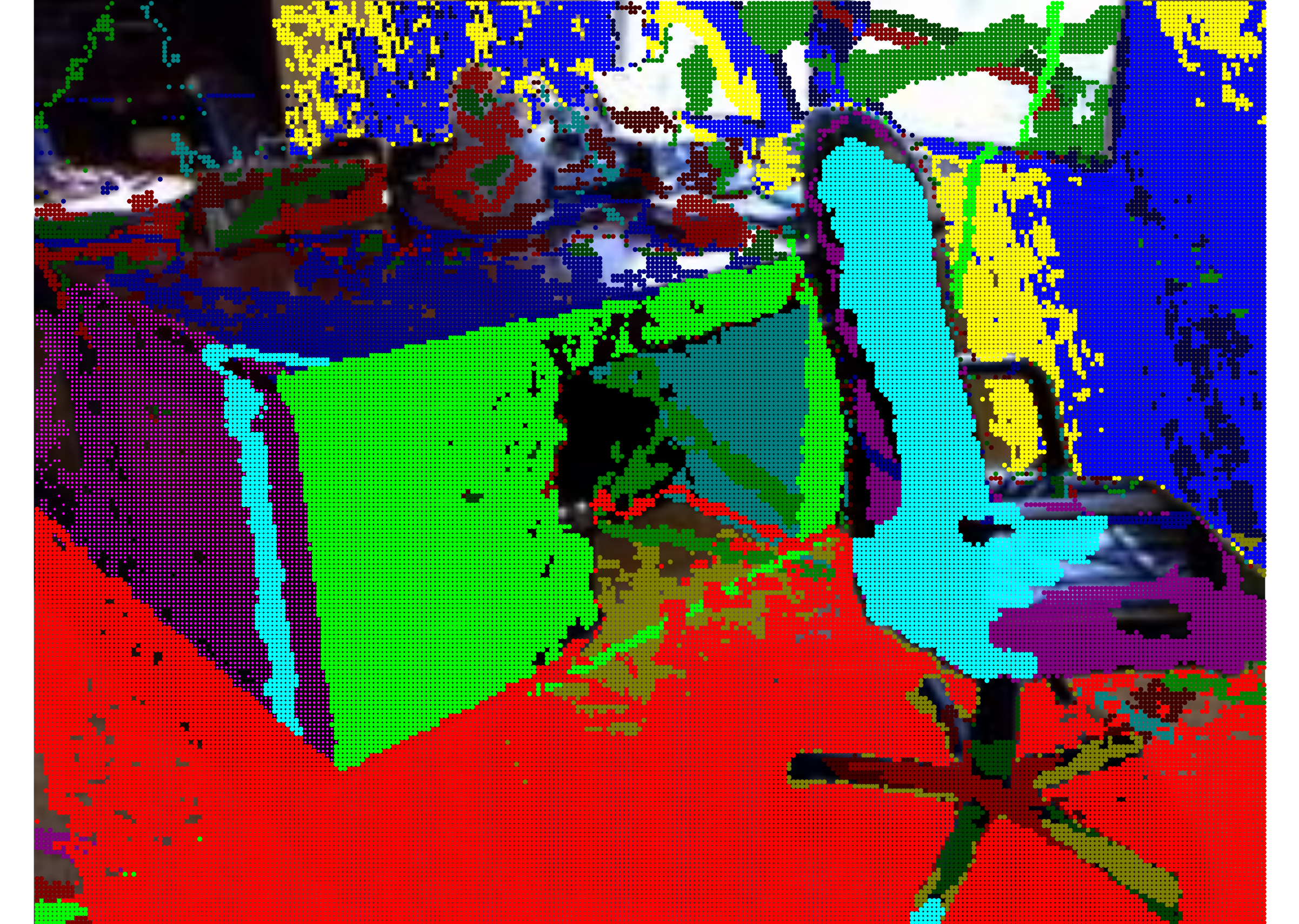} &
    \includegraphics[width=0.24\textwidth]{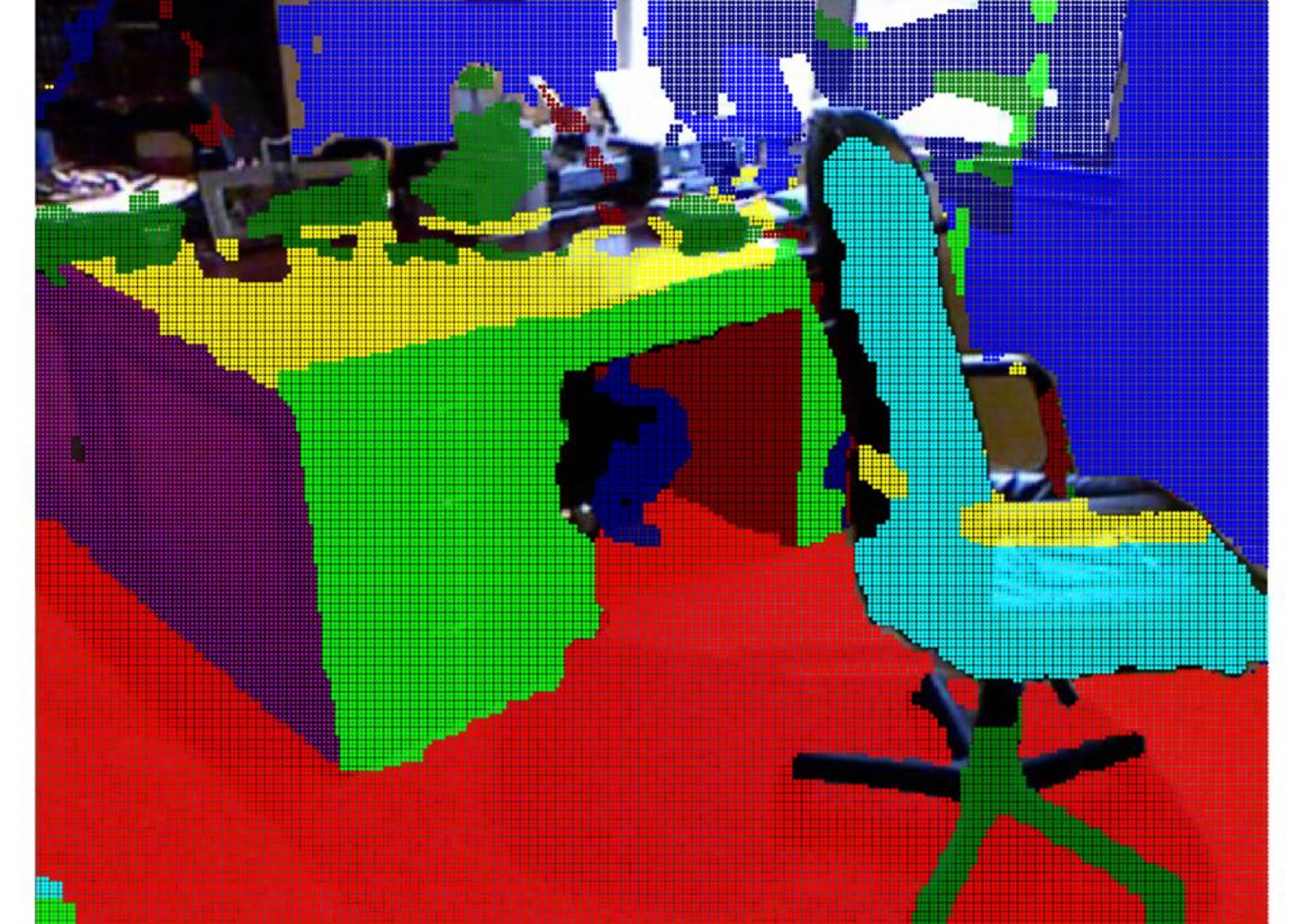} &
    \includegraphics[width=0.24\textwidth]{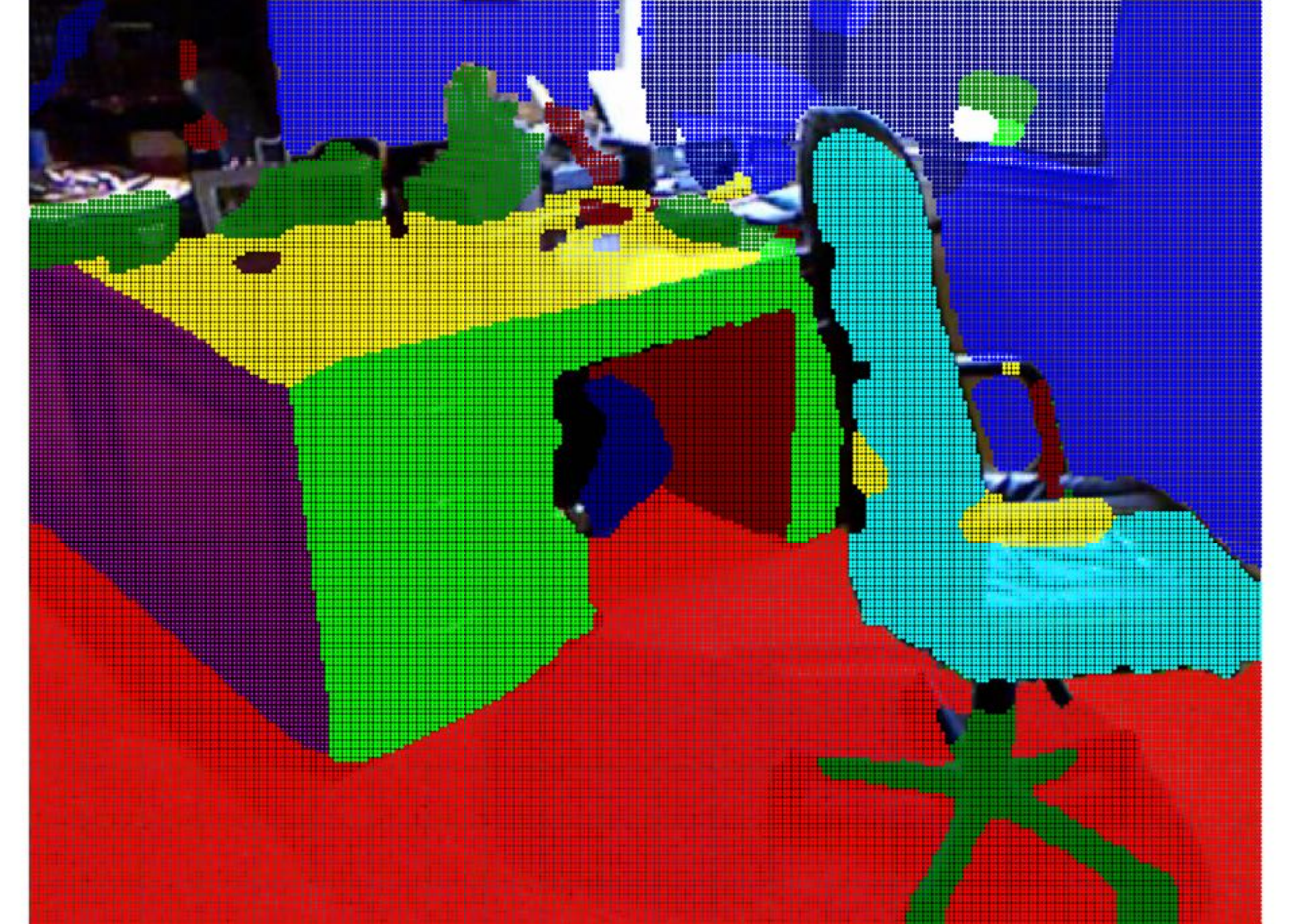} \\
        
    \includegraphics[width=0.24\textwidth]{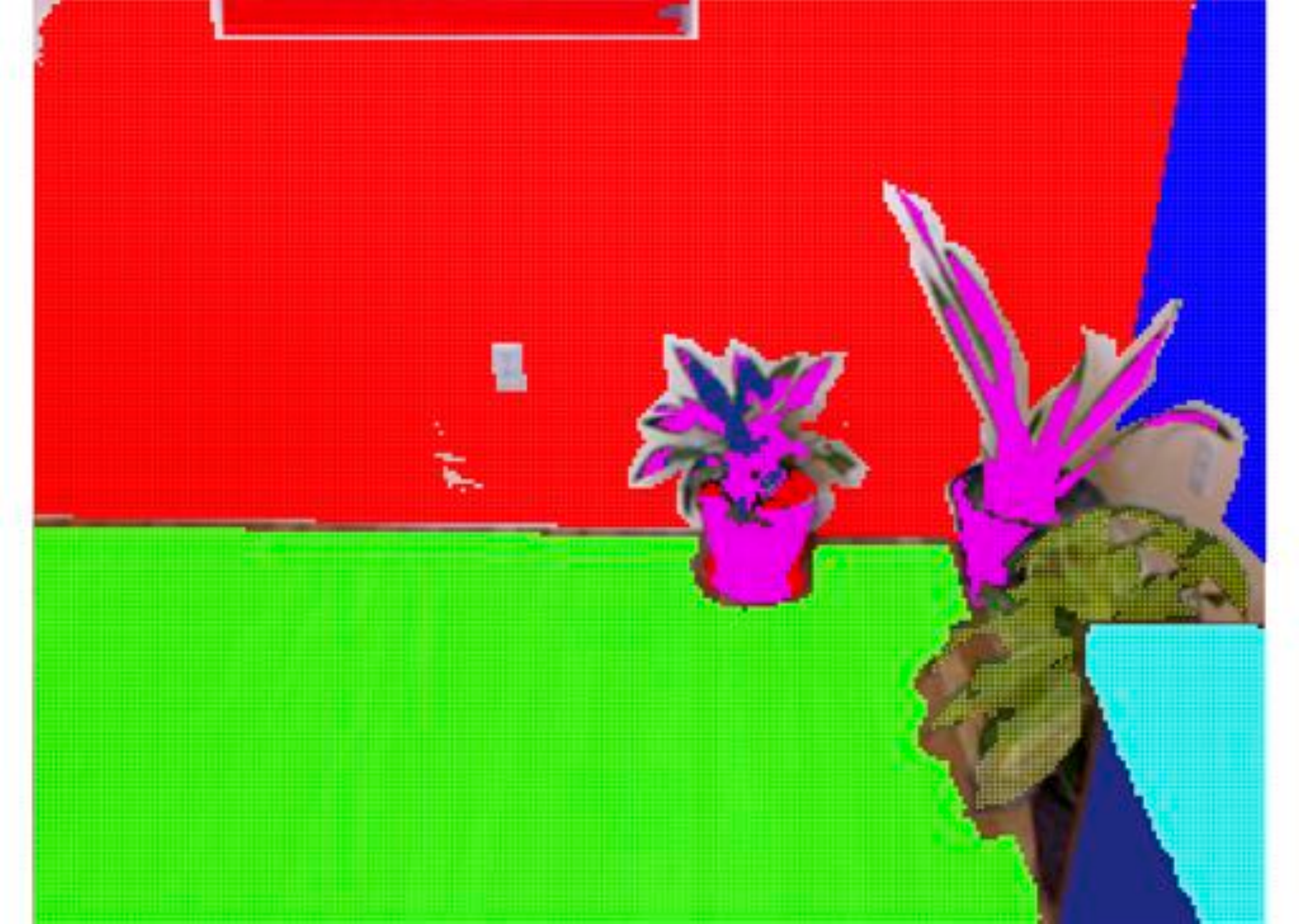} &
    \includegraphics[width=0.24\textwidth]{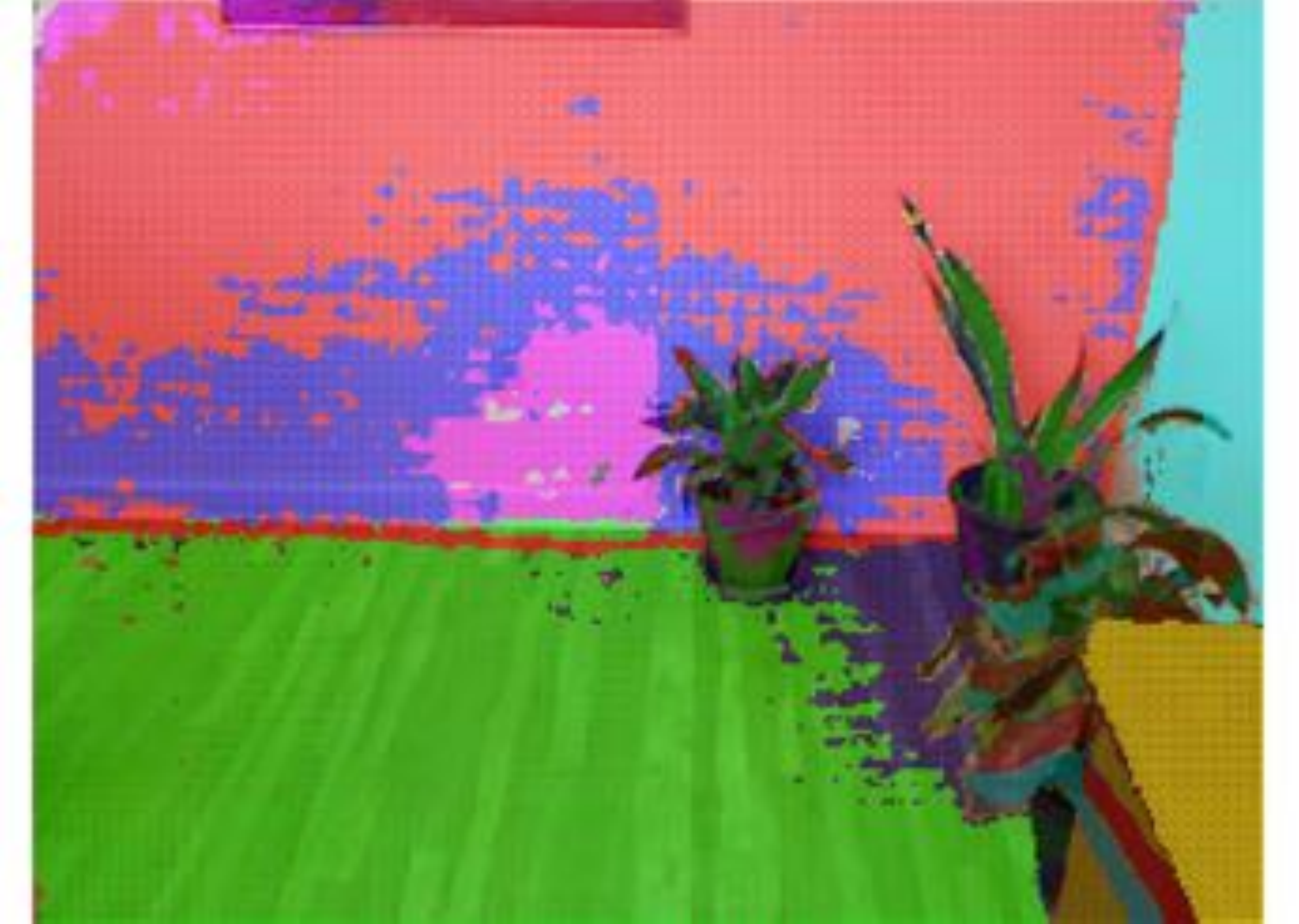} &
    \includegraphics[width=0.24\textwidth]{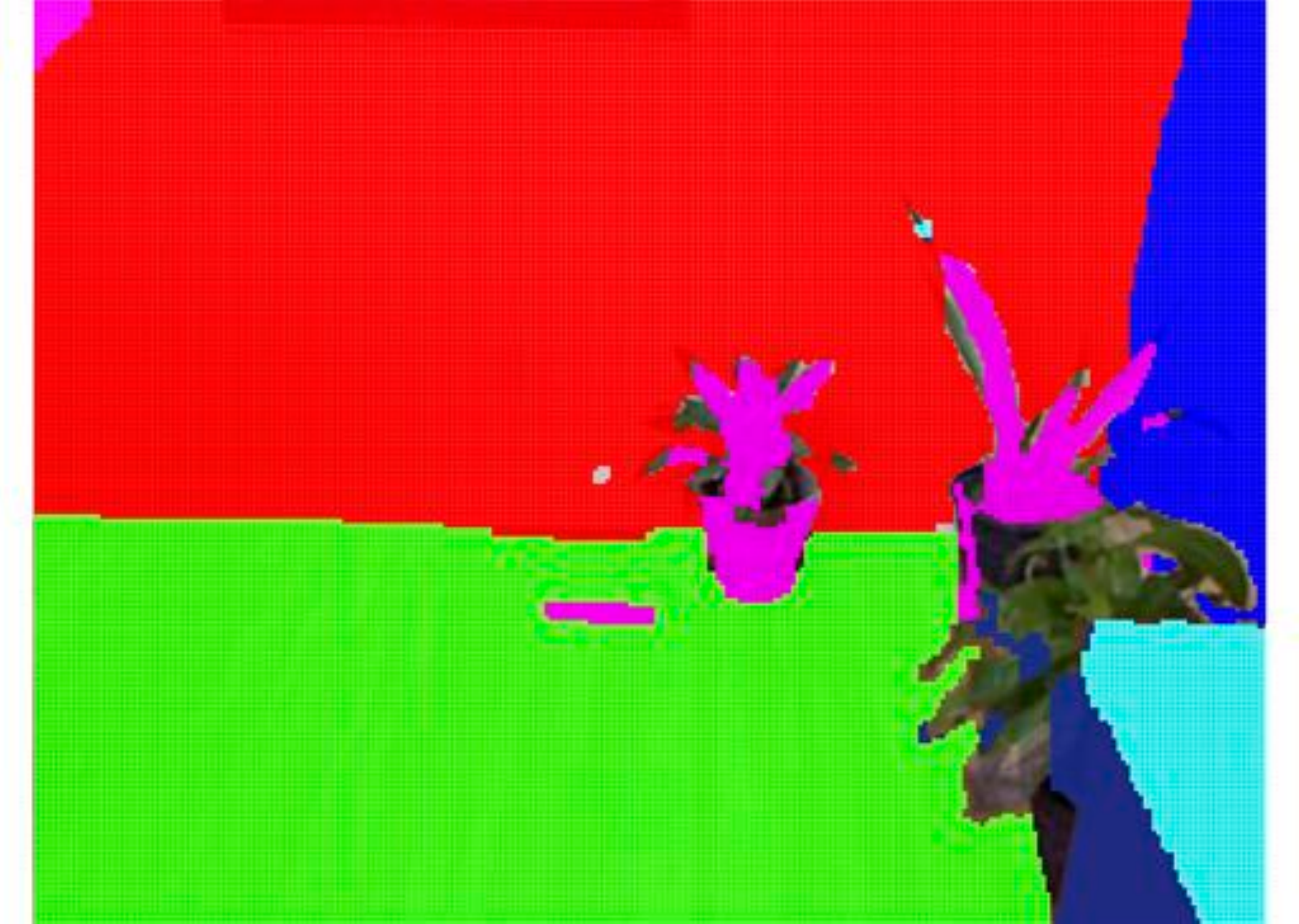} & 
    \includegraphics[width=0.24\textwidth]{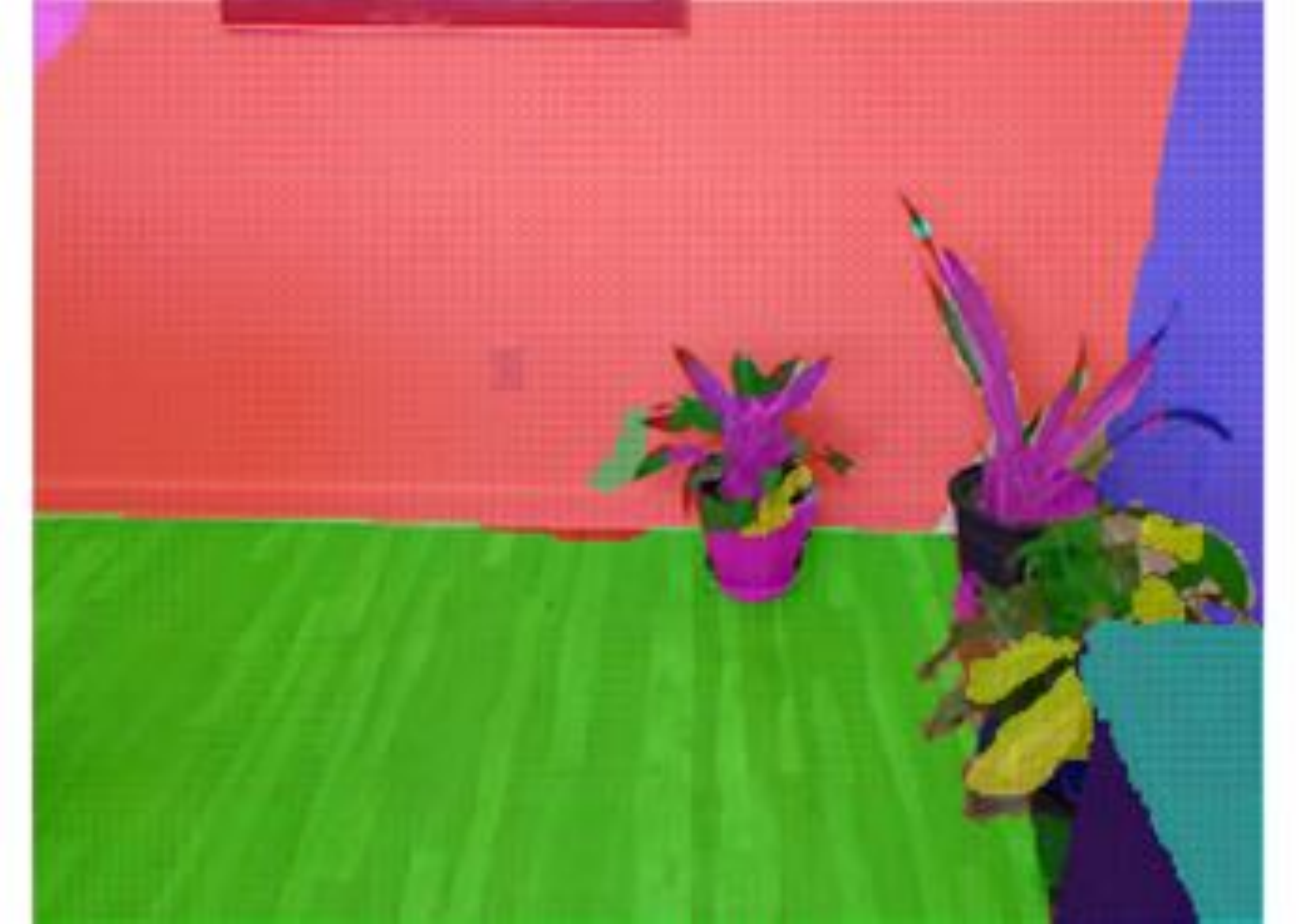} \\

    \end{tabular}
        \caption{Plane segmentation on a sample of RGB-D images from the NYU-Depth dataset. with the Ground Truth, RANSAC, PEARL and CORAL results presented for each image. Pixel membership to a plane model is shown by superposing on a pixel a color-coded point that assigns it to a specific plane model.}
    \label{fig:NYUsegmentation}
\end{table*}

To evaluate the performance of CORAL for plane detection using RGB-D images, we use the NYU-depth dataset \cite{Silberman:ECCV12}. This contains 1449 tuples of RGB, depth and labelled images for multiple instances of objects. For our evaluation, we selected a subset of 232 images containing scenes where either the walls, ceilings, desks,  floor or a combination of all of these surfaces are observed. This is to ensure that our selection contains images with significant planar regions. This subset however was not explicitly created for plane extraction. Therefore a suitable ground truth for our problem is not available. To obtain a ground truth, we employ the labels provided and fit planes to individual instances of objects in the scene with a minimum number of inliers. In addition, we merge planes with similar planar parameters to reduce redundancy of models in the data. 

Four samples of the dataset are shown in Table \ref{fig:NYUsegmentation}. Each column represents the ground truth, RANSAC, PEARL and CORAL solutions with a color-code representing the different labels. The results show that the energy based methods produce more consistent plane models aligned with the expected ground truth. CORAL solutions show greater quality compared to PEARL solutions, in particular considering the first two images. In such cases, CORAL is able to identify more plane models than PEARL.

Table \ref{tab:BenchmarkME_RGBD} condenses the ME results for all three methods. As it is observed, CORAL outperforms PEARL in three of the four instances. To quantify performance over the whole subset used, an optimal value of $\lambda$ was trained using a quarter of the images in the subset. This was then evaluated over the remaining images with the mean errors over all three methods reported back in the Table. 

\begin{table}[h]
    \centering
    \begin{tabular}{ |c| c| c| c|}
    \hline
          & RANSAC & PEARL & \textbf{CORAL}  \\ \hline
          Image 1  & 22.96 & 16.24 & \textbf{13.95}  \\
          Image 2  & 28.70 & 20.59 & \textbf{17.12}  \\ 
          Image 3  & 36.60 & 26.30 & \textbf{25.30}  \\ 
          Image 4  & 15.72 & \textbf{7.77} & 7.83  \\ \hline
           Test set  & 29.38 & 23.04 & \textbf{18.99}  \\ \hline
    \end{tabular}
    \caption{ME for RGB-D plane segmentation on the images shown in Table \ref{fig:NYUsegmentation} and a test set consisting of 232 images from the NYU-depth Dataset. }
    \label{tab:BenchmarkME_RGBD}
\end{table}


 


\section{Discussion and Conclusions}
\label{sec:conclusions}

In this work we have introduced CORAL, a global energy minimisation algorithm for geometric multi-model fitting. Our general solution uses a convex relaxation for model assignment leveraging advanced optimisation techniques in the continuous domain. Our approach inherits all the benefits described in the available survey presented in \cite{nieuwenhuis-et-al-ijcv13}. The advantages of CORAL over discrete solutions analysed in previous sections are mainly driven by its potential to parallel implementation compared to state-of-the-art methods that require sequential evaluations of labels. CORAL intrinsically boosts performance by simultaneously handling per-point evaluations over the labels. Additionally this speed-up comes bundled with reduced runtime variance as compared to graph-cut approaches. Making our approach more suitable for geometric model extraction with applications with real-time performance constraints.

Our formulation also allows for flexibility on the norms applied, which we characterise through the structure detection problem from images in two different scenarios. All this without degradation of performance as in these scenarios CORAL reports results that outperform or are at least equivalent to the best performances achieved using state-of-the-art methods. 

In summary, CORAL brings in powerful optimisation machinery into the solution of geometric multi-model fitting. Offering an algorithm that is simultaneously able to robustly extract accurate models in the presence of contamination and offers improved time performance guarantees over the state of the art.





\bibliography{coral.bbl}
\bibliographystyle{IEEEtranS}

\end{document}